\documentclass[11pt,a4paper]{article}
\usepackage[hyperref]{arxiv}
\usepackage{times}
\usepackage{latexsym}
\usepackage{comment}

\usepackage{microtype}

\usepackage{graphicx} 
\usepackage{amsmath}
\usepackage{wrapfig,booktabs}
\usepackage{multirow}
\usepackage{bbm}
\usepackage{wrapfig}
\usepackage{subfigure}

\usepackage{graphicx}
\usepackage{amsmath}
\usepackage{booktabs}
\usepackage{algorithm}
\usepackage{algorithmic}
\usepackage{color}
\usepackage{multirow}
\usepackage{graphbox}
\usepackage{theorem}
\newtheorem{prop}{Proposition}

\newtheorem{definition}{Definition}

\usepackage{subfigure}
\usepackage{framed}
\usepackage{soul}
\usepackage{url}
\usepackage{amssymb}
\usepackage{bm}
\aclfinalcopy

\def\shownotes{0}  
\ifnum\shownotes=1
\newcommand{\authnote}[2]{[#1: #2]}
\else
\newcommand{\authnote}[2]{}
\fi

\title{Don't Touch What Matters: Task-Aware Lipschitz Data Augmentation  \\ for Visual Reinforcement Learning}

\author{
Zhecheng Yuan$^1$,
Guozheng Ma$^1$,
Yao Mu$^{2}$,
Bo Xia$^1$, 
Bo Yuan$^1$,\\
\textbf{Xueqian Wang$^1$,
Ping Luo$^{2}$,
Huazhe Xu$^{3}$} \\
$^1$Tsinghua University $^2$Hongkong University $^3$Standford University \\
\texttt{yuanzc20@mails.tsinghua.edu.cn,
huazhexu@stanford.edu}
}

\begin{document}
\maketitle

\begin{abstract}
One of the key challenges in visual Reinforcement Learning (RL) is to learn policies that can generalize to unseen environments. Recently, data augmentation techniques aiming at enhancing data diversity have demonstrated proven performance in improving the generalization ability of learned policies. However, due to the sensitivity of RL training, naively applying data augmentation, which transforms each pixel in a task-agnostic manner, may suffer from instability and damage the sample efficiency, thus further exacerbating the generalization performance. At the heart of this phenomenon is the diverged action distribution and high-variance value estimation in the face of augmented images. To alleviate this issue, we propose \textbf{T}ask-aware \textbf{L}ipschitz \textbf{D}ata \textbf{A}ugmentation (TLDA) for visual RL, which explicitly identifies the task-correlated pixels with large Lipschitz constants, and only augments the task-irrelevant pixels. To verify the effectiveness of TLDA, we conduct extensive experiments on DeepMind Control suite, CARLA and DeepMind Manipulation tasks, showing that TLDA improves both sample efficiency in training time and generalization in test time. It outperforms previous state-of-the-art methods across the 3 different visual control benchmarks\footnote{\url{https://sites.google.com/view/algotlda/home}}.

\end{abstract}

\section{Introduction}
Deep Reinforcement Learning (DRL) from visual observations has carved out brilliant paths in many domains such as video games~\cite{mnih2015human}, robotics manipulation~\cite{kalashnikov2018scalable}, and visual navigation~\cite{zhu2017target}. However, it remains challenging to obtain generalizable policies across different environments with visual variations due to overfitting~\cite{zhang2018study}.

\begin{figure}[t]
  \centering
  \includegraphics[width=0.9\columnwidth]{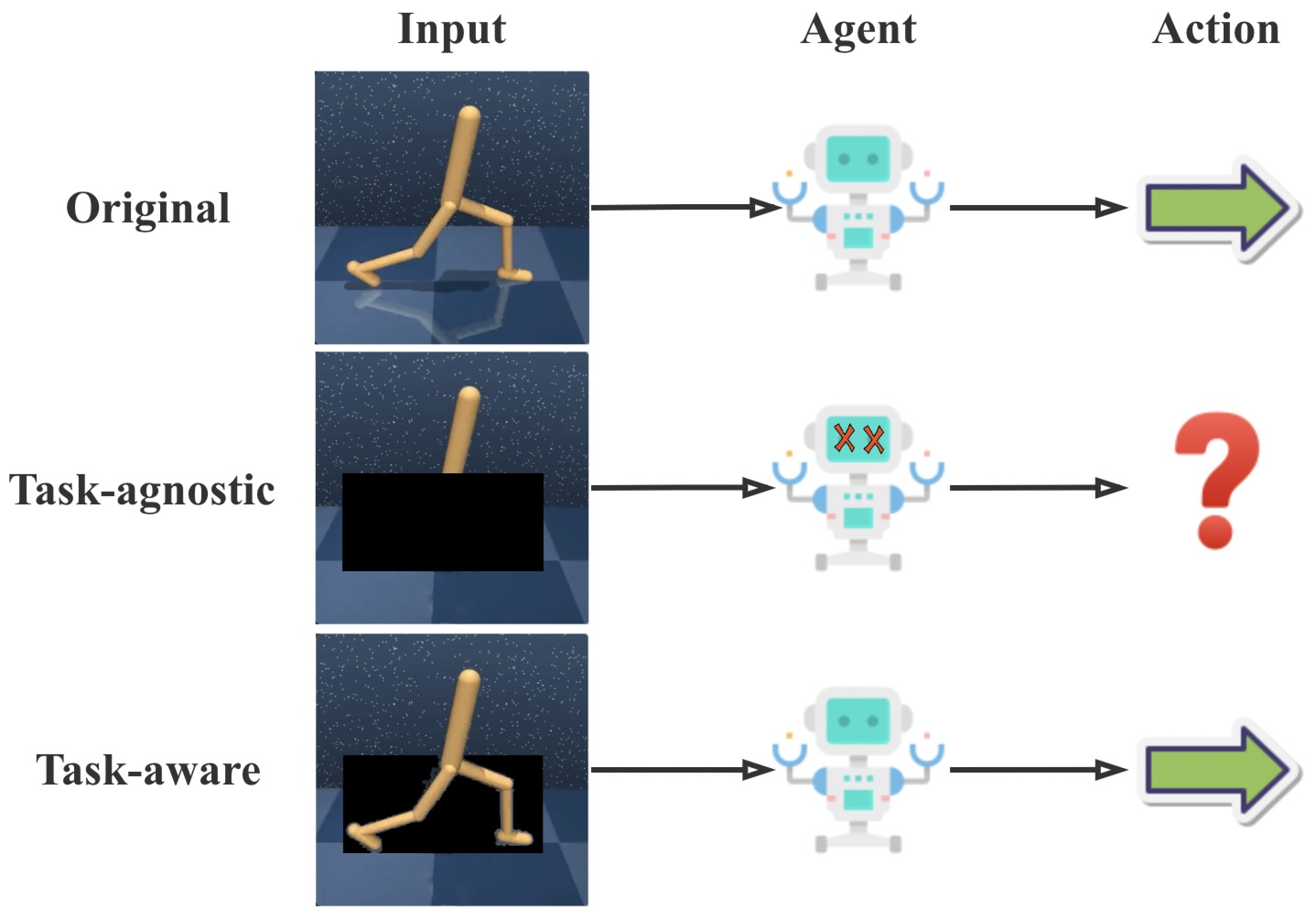}
   \vspace{-5pt}
   \caption{Augmenting observation in a task-agnostic manner (in the \textit{middle}) distracts the agent's decision, hence it will damage agent's asymptotic performance. This problem can be alleviated by task-aware data augmentation (in the \textit{bottom}).}
   \label{fig:semantic}
\vspace{-20pt}
\end{figure}

Data Augmentation~\cite{shorten2019survey} and Domain Randomization~\cite{tobin2017domain}  based approaches are widely used to learn generalizable visual representations. However, recent work~\cite{hansen2021stabilizing} find that in visual RL, there is a dilemma: heavy data augmentations are vital for better generalization, but it will cause a significant decrease in both the sample efficiency and the training stability. 

\begin{figure*}[!t]
    \vspace{-15pt}
    \centering
    \setlength{\abovecaptionskip}{5pt}
    \subfigure[ \label{fig:action-error-strong}]
        {\includegraphics[width=0.295\linewidth]{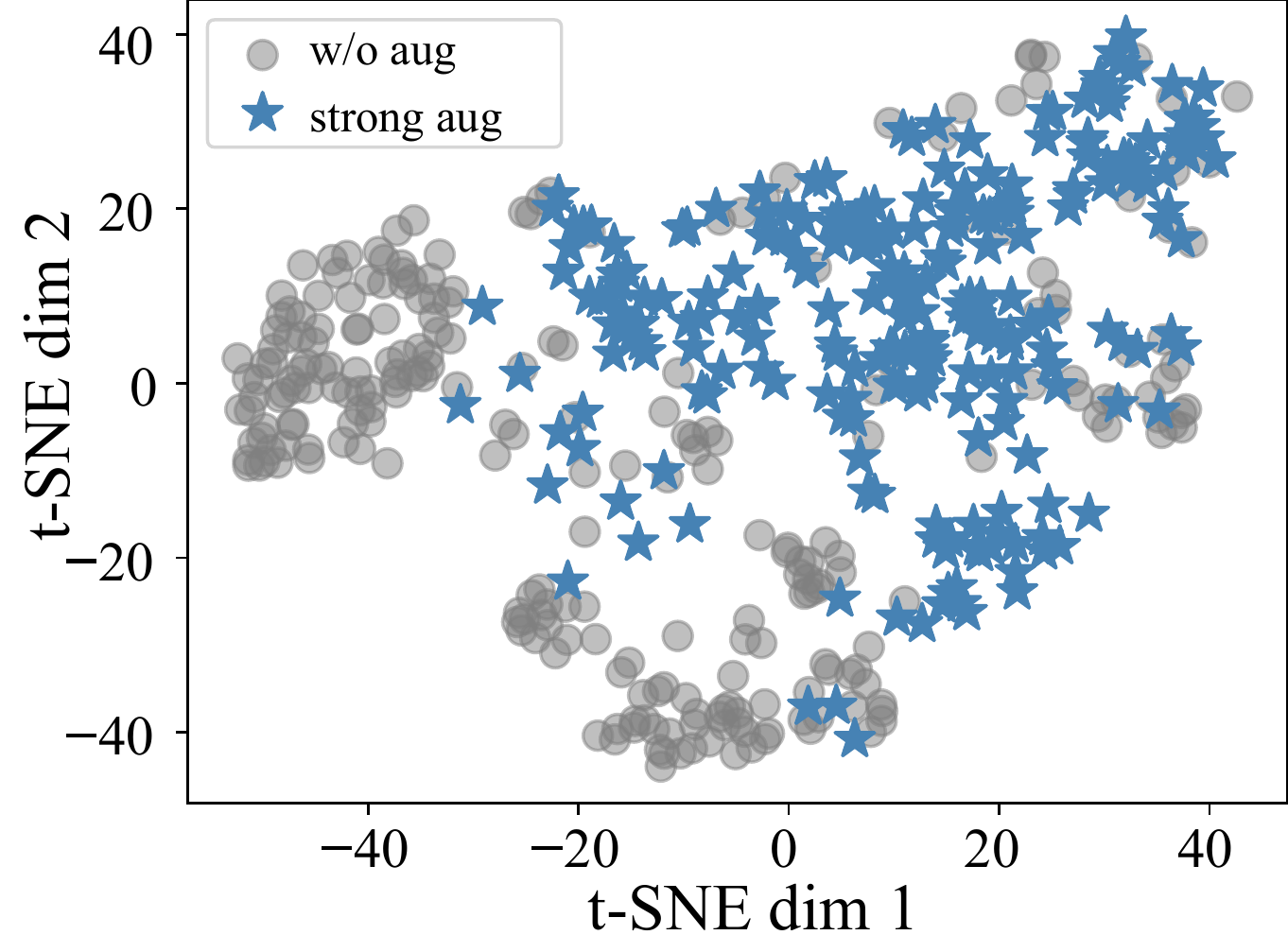}}
    \hspace{6mm}
    \subfigure[ \label{fig:action-error-ours}]
        {\includegraphics[width=0.295\linewidth]{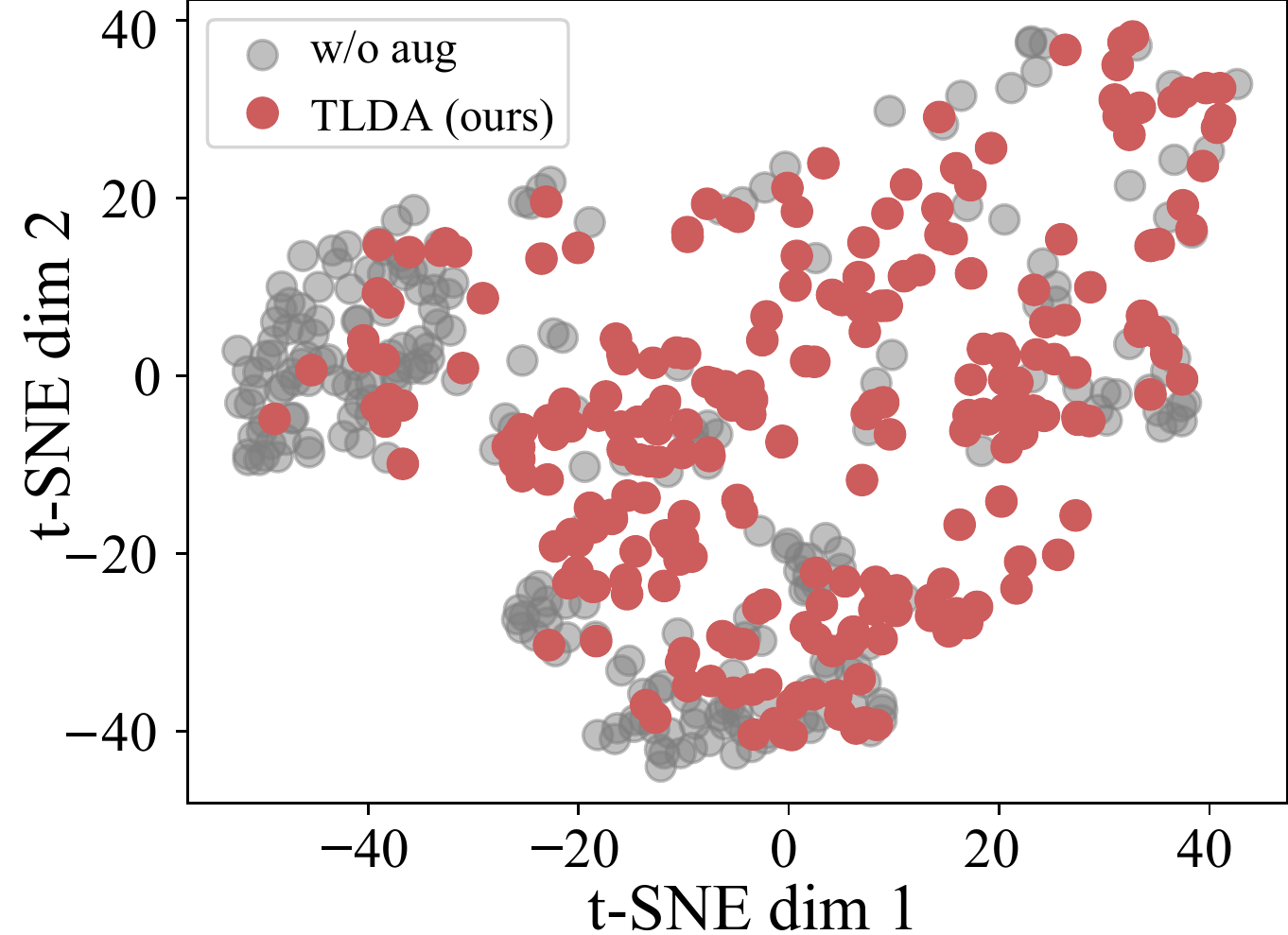}}
    \hspace{6mm}
    \subfigure[ \label{fig:action-error-weak}]
        {\includegraphics[width=0.295\linewidth]{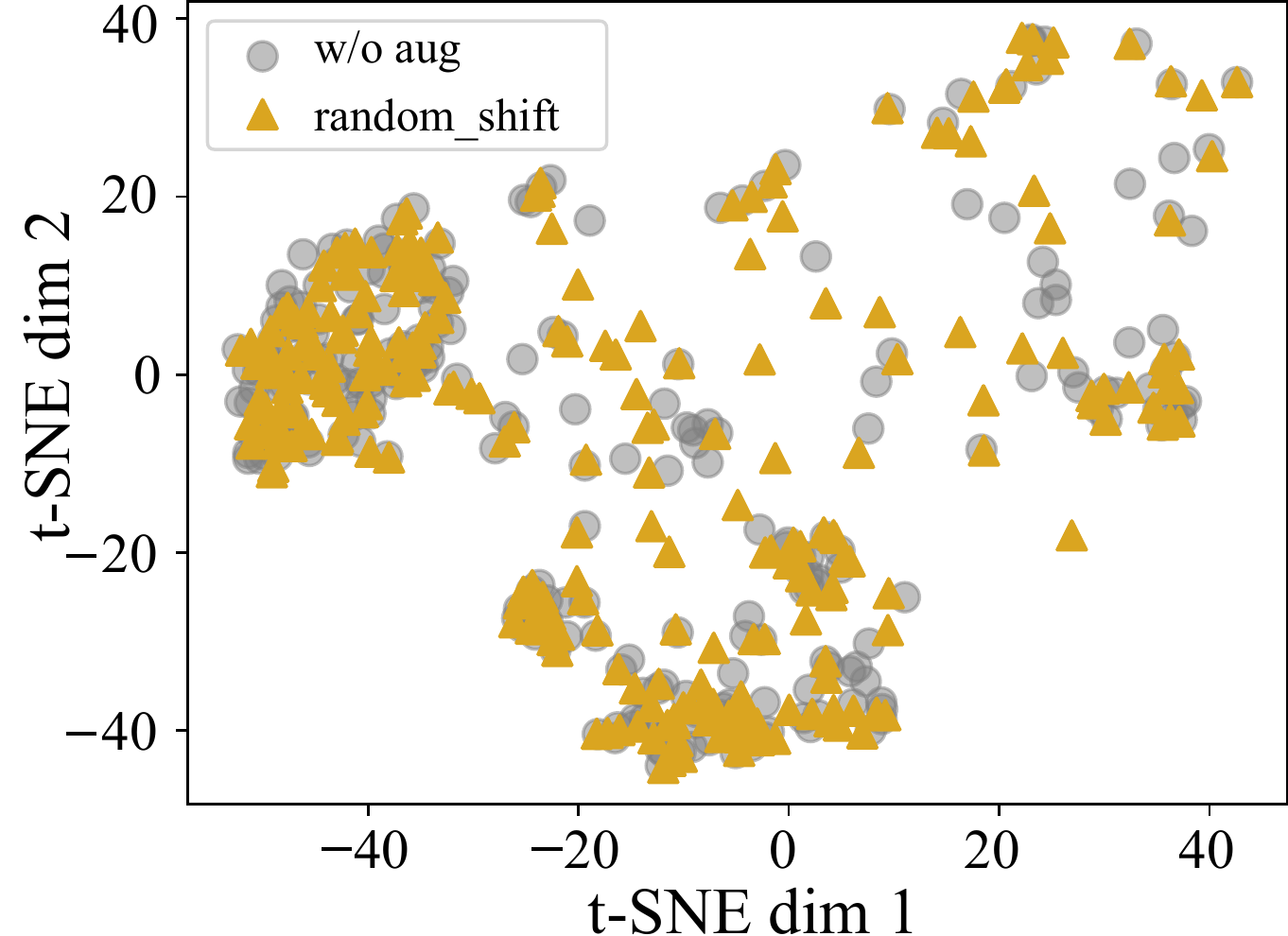}}
\vspace{-5pt}
\caption{\textbf{Action Distribution.}  We use t-SNE to show high-dimensional actions employed by the same agent. The \textbf{grey dots} are the actions given the observations without augmentation (\textit{w/o aug}); \textbf{the blue dots} (a) and \textbf{orange dots} (c) are the actions given the same observations under strong (\textit{random\_conv}) and weak augmentation (\textit{random\_shift}), respectively. The visualized results demonstrate that there is a significant action distribution shift under strong augmentation, while under weak augmentation, the policy is closer to the original distribution. The \textbf{red dots} (b) are TLDA under strong augmentation, which gives rise to an action distribution that remains similar to \textbf{the grey dots}.}
\label{fig1}
\vspace{-10pt}
\end{figure*}

One of the main reasons is that data augmentation conventionally perform pixel-level image transformation, where each pixel is transformed in a task-agnostic manner. However, in visual RL, each pixel in the observation has different relevance to the task and the reward function. Hence, it is worth rethinking data augmentation in the new context of visual RL.

To better understand the effect of data augmentation in visual RL, we visualize the action distribution output from policies trained with various data augmentation choices in Figure~\ref{fig1}. We find that the agent's actions vary dramatically when faced with different augmentation methods. Specifically, when weak augmentation such as \textit{shifting} is applied, the action distribution remains closer to the original distribution that has no augmentation (Figure~\ref{fig:action-error-weak}); however, when strong augmentation e.g., \textit{random convolution} is applied, the action distribution drastically changes (Figure~\ref{fig:action-error-strong}) and the Q-estimation yields the  discrepancy with the un-augmented data, as shown in Figure~\ref{fig2}. This comparison reveals the severe problem that causes instability when data augmentation is applied blindly without knowing the task information.

In this work, we propose a task-aware data augmentation method in visual RL that learns to augment the pixels less correlated to the task, namely \textbf{T}ask-aware  \textbf{L}ipschitz \textbf{D}ata \textbf{A}ugmentation (TLDA), as shown in Figure~\ref{fig:semantic}. A desirable quality for such a method is to that it maintains a stable policy output even on augmented observations. Following this insight, we introduce the \textit{Lipschitz constant} that measures the relevance between the pixel and the task, then guides the augmentation strategy. Specifically, we first impose a perturbation on a certain pixel, and calculate the corresponding Lipschitz constant for the pixel via the policy change before and after the perturbation. Then, to avoid the occurrence of drastic policy changes, we treat the pixels with larger Lipschitz constant as the task-correlated ones and avoid augmenting them. Therefore, the output could be more stable while keeping the diversity of augmented data.

We conduct experiments on 3 benchmarks: DMControl Generalization Benchmark (DMC-GB)~\cite{hansen2021generalization}, CARLA~\cite{dosovitskiy2017carla}, and DMControl manipulation tasks~\cite{tunyasuvunakool2020dm_control}. We train agents in a fixed environment and evaluate their generalization on the environments that are unseen during training. Extensive experiments show that TLDA  outperforms the prior state-of-the-art methods due to more stable and efficient training and robust generalization performance.

Our main contributions are summarized as follows:
\vspace{-2pt}
\begin{itemize}
\vspace{-4pt}
    \item We propose \textbf{T}ask-aware \textbf{L}ipschitz \textbf{D}ata \textbf{A}ugmentation~(TLDA), which can be implemented on any downstream visual RL algorithm easily without adding auxiliary objectives or additional learnable parameters.
    \vspace{-4pt}
    \item We provide theoretical understanding and empirical results to show TLDA can alleviate the action distribution shift and high variance Q-estimation problems effectively. 
    \vspace{-4pt}
    \item TLDA achieves competitive or better sample efficiency and generalization ability than previous state-of-the-art methods in $3$ different kinds of benchmarks.

\end{itemize}

\section{Related Work}
\textbf{Generalization in RL}. Researchers have been investigated RL generalization from various perspectives, such as different visual appearances~\cite{cobbe2019quantifying}, dynamics~\cite{packer2018assessing} and environment structures~\cite{cobbe2020leveraging}. In this paper, we focus on generalization over different visual appearances. Two popular paradigms are proposed to address the overfitting issue in current visual RL research. The first is to regard generalization as a representation learning problem. For example, Bi-simulation metric~\cite{ferns2011bisimulation} is implemented to learn robust representation features~\cite{zhang2020learning}. The other paradigm is to design auxiliary tasks. SODA~\cite{hansen2021generalization} adds a BYOL-like~\cite{grill2020bootstrap} architecture and introduces an auxiliary loss which encourages the representation to be invariant to task-irrelevant properties of the environment. In contrast to the previous efforts, our method does not require to employ a specific metric to learn representation, nor to introduce additional modules.

\textbf{Data Augmentation for RL}. Data Augmentation is an efficient method to improve the generalization of visual RL. RAD~\cite{laskin2020reinforcement} compares different data augmentation methods and reveals that the benefits of different augmentation methods to RL tasks are not the same. SECANT~\cite{fan2021secant} mentions that weak augmentation can improve sample efficiency but not generalization ability. It also shows that the simple use of strong augmentation is likely to cause training divergence, though generalization ability is improved. Automatic data augmentation is proposed in~\cite{raileanu2020automatic} to make better use of data augmentation. We advocate this paradigm and believe that one crucial factor for improving sample efficiency and generalization lies in the design of data augmentation, namely, how we can diversify the input as much as possible while maintaining the invariance of output. We show that how strong augmentation affects action distribution shifts and causes high variance of Q estimation, and illustrate that our approach is effective in alleviating these two problems.

\section{Preliminaries}
We consider learning in a Markov Decision Process (MDP) formulated by the tuple $\langle \mathcal { S } ,  \mathcal { A } , r , \mathcal{P} ,  \gamma \rangle$ 
where $\mathcal{S}$ is the state space, $\mathcal{A}$ is the action space, $r:\mathcal{S} \times \mathcal{A} \mapsto \mathbb{R}$ is a reward function, $\mathcal{P}\left({s}_{t+1} \mid {s_{t}}, {a_{t}}\right)$ is the state transition 
function, $\gamma \in[0,1)$ is the discount factor. The goal is to learn a policy ${\pi^{*}}$ to maximize the expected cumulative return  $\pi^{*} = \operatorname{argmax}_{\pi} \mathbb{E}_{a_{t} \sim \pi\left(\cdot \mid s_{t}\right),s_{t} \sim \mathcal{P}}\left[\sum_{t=1}^{T} \gamma^{t} r\left(s_{t}, a_{t}\right)\right]$, starting from an initial state ${s}_{0} \in \mathcal{S}$ and following the policy $\pi_{\theta}\left(\cdot \mid s_{t}\right)$ which is parameterized by a set of learnable parameters $\theta$. Meanwhile, we expect the learned policy $\pi^{*}_{\theta}$ can be well generalized to new environments, which have the same structure and definition of the original MDP, but with different observation space $\mathcal{O}$ constructed from the same state space $\mathcal{S}$.

\begin{figure*}[t]
  \vspace{-15pt}
  \centering
  \includegraphics[width=0.95\linewidth]{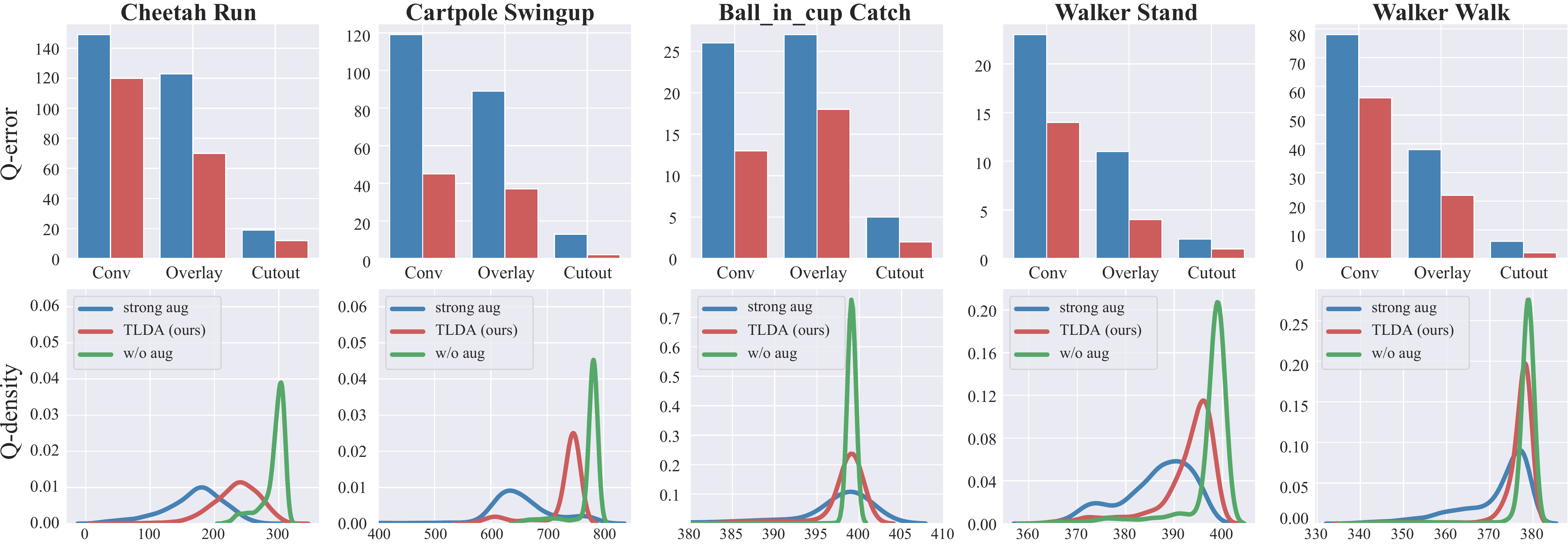}
     \vspace{-5pt}
   \caption{\textbf{Q-estimation error}. \textit{Top:} We measure the Q-estimation mean square error of the different augmented observation vs.the non-augmented observation. \textbf{The blue bar} and \textbf{the red bar} are the error between strong augmented data and TLDA-augmented data vs non-augmented data, separately. It shows that TLDA can significantly reduce the Q-estimation error to alleviate the high-variance estimation problems. \textit{Bottom:} The distribution of Q-estimation. TLDA comes up with a closer Q-estimation distribution with the original one.}
   \label{fig2}
   \vspace{-10pt}
\end{figure*}

\subsection{Data Augmentation}

\begin{definition} \label{d1}
\textbf{(Optimality-Invariant State Transformation)} \\
Given an MDP $\mathcal{M}$, we define an augmentation method $\phi: \mathcal{O} \rightarrow \mathcal{O}^{\prime}$ as an optimality-invariant transformation if $\forall o \in \mathcal{O}, a \in \mathcal{A}, \phi(o) \in \mathcal{O'} $,  where $\mathcal{O'}$ is a new set of observation satisfies:
\begin{gather}
Q(o, a)=Q(\phi(o), a) \quad \pi(\cdot \mid o)=\pi(\cdot \mid \phi(o)) 
\end{gather}
\end{definition}

A desirable quality for data augmentation is to satisfy the form of Optimality-Invariant State Transformation while distortion or distracting noise is added to the observation.

\subsection{Lipschitz constant} \label{subection:Lipschitz constant}
The Lipschitz constant is frequently utilized to measure the robustness of a model, we introduce Lipschitz continuity of the policy here. A function $f$: $\mathbb{R}^{n} \rightarrow \mathbb{R}^{m}$ is Lipschitz continuous on $\mathcal{X} \subseteq \mathbb{R}^{n}$ if there exists a non-negative constant $K \geq 0$ such that
\begin{equation}
\|f(x)-f(y)\| \leq K\|x-y\| \text { for all } x, y \in \mathcal{X}
\end{equation}

The smallest such $K$ is called the Lipschitz constant of $f$~\cite{pauli2021training}.

\begin{definition}[\textit{Lipschitz constant of the policy}] \label{d2}
Assume the state space is equipped with a distance metric $d(\cdot , \cdot)$. Under a certain augmentation method $\phi$, the Lipschitz constant of a policy $\pi$ is defined as follows:

\begin{equation}
\label{defff}
K _ { \pi } = \sup _ { s \in \mathcal { S } } \frac { D _ { T V } \left( \pi \left( \cdot \mid \phi ( s ) \right) \| \pi \left( \cdot \mid  s  \right) \right) } { d(\phi(s), s)}
\end{equation}
where $D_{TV}(P||Q) = \frac{1}{2} \sum_{a \in \mathcal{A}}|P(a)-Q(a)|$ is the total variation distance between distributions. If $K _ { \pi }$ is finite, the policy $\pi$ is Lipschitz continuous. 

\end{definition}

For a certain model, a smaller Lipschitz constant generally represents higher stability against the variance of input ~\cite{finlay2018lipschitz}. The following proposition illustrates that the estimation error of Q-value can be bounded by the Lipschitz constant:

\vspace{-5pt}
\begin{prop}
We consider an MDP $\mathcal{M}$, a policy $\pi$ and an augmentation method $\phi$. Suppose the rewards are bounded by $r_{max}$ and state space is equipped with a distance metric $d(\cdot , \cdot)$, such that $\forall a \in \mathcal{A}, \forall s \in \mathcal{S}, |r(s, a)| \leq r_{max}$, the following inequality holds, where $\left\| d(\phi) \right\| _ { \infty }= \sup _ { s \in \mathcal { S } } d(\phi(s), s):$
\begin{equation}
\begin{small}
\begin{aligned}
\left| Q ^ { \pi } ( s , a ) - Q ^ { \pi } ( \phi ( s ) , a ) \right| \leq 2r_{max}\frac{(K_{\pi}\left\| d(\phi) \right\| _ { \infty } + 1)}{1-\gamma}
\end{aligned}
\end{small}
\end{equation}
\end{prop}

The formal statement and the proof are shown in Appendix~\ref{a1}. This proposition indicates that if a smaller Lipschitz constant under one specific augmentation is acquired, we will have a tighter bound of the Q-value estimation with a lower variance while implementing data augmentation.

\section{Method}
To maintain the training stability and improve the generalization ability, we propose: \textbf{T}ask-aware \textbf{L}ipschitz \textbf{D}ata \textbf{A}ugmentation (TLDA), an efficient and general task-aware data augmentation method for visual RL.

\begin{figure*}[t]
  \centering
  \vspace{-10pt}
  \includegraphics[width=0.97\linewidth]{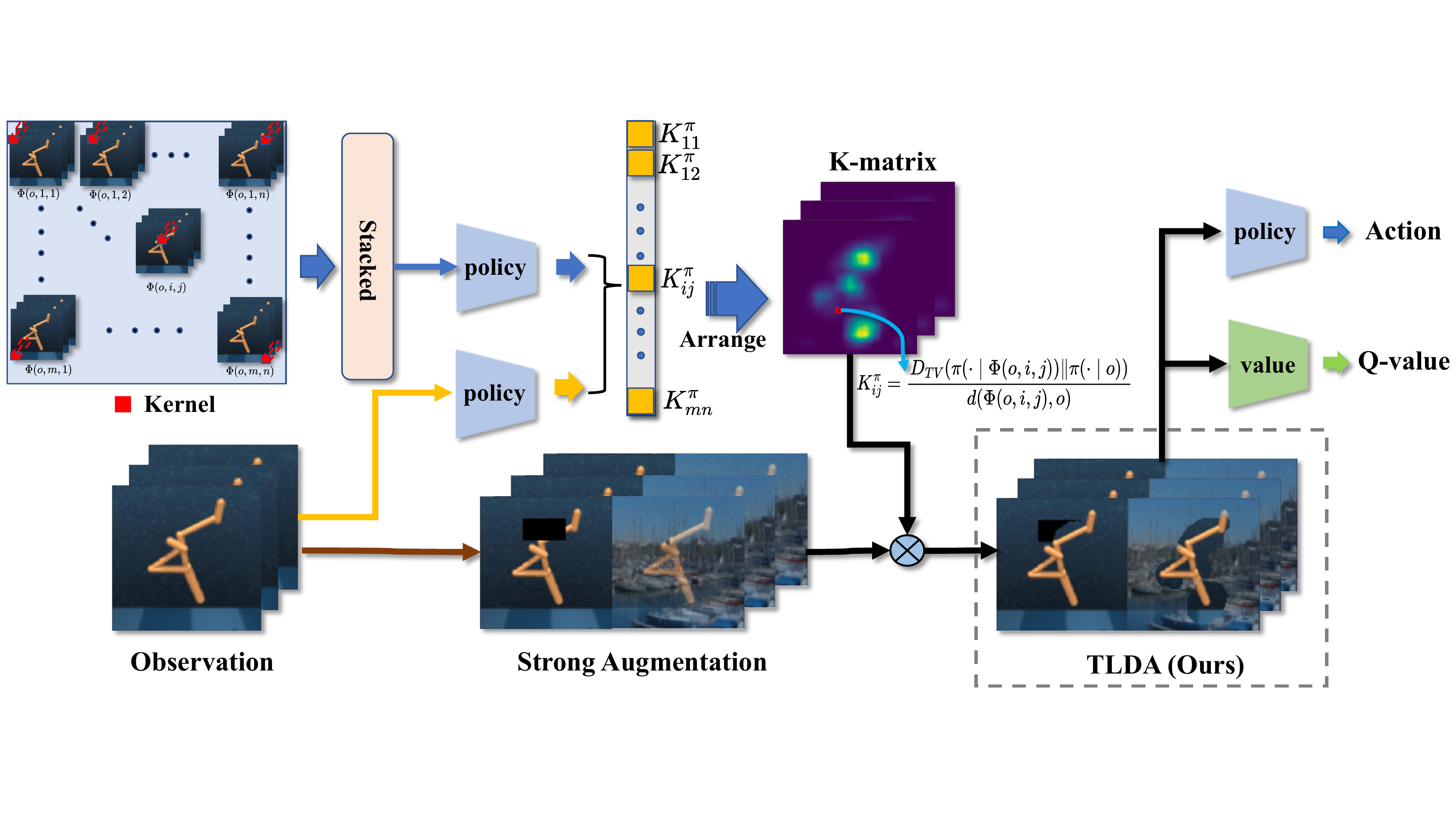}
  \vspace{-11pt}
   \caption{\textbf{Overview of TLDA}. This figure shows two examples of TLDA and the pipeline implementing it. The agent generates the \textit{K-matrix} for a  stacked frame, and then preserves the larger Lipschitz constant areas under strong augmentation. The preserved areas are highlighted in the \textit{K-matrix}. }
   \label{fig3}
     \vspace{-10pt}
\end{figure*}

\subsection{Construct the \textit{K-matrix}}
\label{section:Construct K}

We first calculate the Lipschitz constant from perturbed input images. By using a kernel to perturb the original image $o \in \mathbb{R}^{H \times W}$, we obtain the perturbed image denoted as $A(o)$. Next, we choose the pixels centered with the location $(i,j)$ of $A(o)$ as in the Eq (\ref{eq9}), denoted as $\Phi(o,i,j)$. Specifically, we use the Hadamard product $\odot$ to choose the perturbed pixels around location$(i, j)$ by an image mask $M(i, j) \in(0,1)^{H \times W}:$ 

\begin{equation} 
\Phi ( o , i , j ) = o \odot ( 1 - M ( i , j ) ) + A \left( o \right) \odot M ( i , j )
\label{eq9}
\end{equation}

To derive the Lipschitz constant, we use the notation $d(\Phi(o,i,j), o)$ to represent the distance between input $o$ and $\Phi ( o , i , j )$ under metric $d(\cdot, \cdot)$. As in Definition~\ref{d2}, for a given observation $o$, the Lipschitz constant of the pixel $(i, j)$ can be computed as follows:

\begin{equation}
K_{i j}^{\pi} = \frac{D_{T V}\left(\pi\left(\cdot \mid \Phi(o,i,j)\right) \| \pi(\cdot \mid o)\right)}{d(\Phi(o,i,j), o)}
\label{K}
\end{equation}
where the numerator can be interpreted as distance between two action distributions: $\pi(\cdot \mid \Phi(o,i,j))$, $\pi(\cdot \mid o)$, and the denominator is the distance between the original observation and the perturbed one.

With the per-pixel Lipschitz constant in hand, we then construct the matrix that can reflect the task-relevance information and be applied on the whole observation.
By arranging $K_{i j}^{\pi}$ into a matrix which have the same size as $o$ following Eq~(\ref{eq100}), we denote this matrix as the \textit{K-matrix}:

\begin{equation}
\begin{small}
\textit{K-matrix} \triangleq \left[ \begin{array} { c c c c } { K } _ { 11 } ^ { \pi } & { K } _ { 12 } ^ { \pi } & \cdots &  { K } _ { 1 n } ^ { \pi } \\  { K } _ { 21 } ^ { \pi } &  { K } _ { 22 } ^ { \pi } & \cdots & { K } _ { 2 n } ^ { \pi } \\ \vdots & \vdots & \ddots & \vdots \\  { K } _ { m 1 } ^ { \pi } &  { K } _ { m 2 } ^ { \pi } & \cdots &  { K } _ { m n } ^ { \pi } \end{array} \right]
\end{small}
\label{eq100}
\end{equation}
We aim to capture the task-related locations with large Lipschitz constants which tend to cause high variance in the policy/value output during the same level of perturbation.

\subsection{Task-Aware Lipschitz Augmentation (TLDA) with the \textit{K-matrix}}

Intuitively, data augmentation operations should not modify the task-related pixels indicated by large Lipschitz constants. We follow this intuition and propose a simple yet effective way to decide which areas can be modified. We use the mean value of the \textit{K-matrix} as a threshold, and binarize the \textit{K-matrix} by the following way, where $N$ is the number of pixels ($H \times W$), $K^{mean} = \frac{1}{N} \times \sum_{ij}K_{ij}^{\pi}:$
\begin{equation}
\begin{split}
\vspace{-10pt}
M_{ij}^{K} = \begin{cases}1, & if \quad K_{ij}^{\pi} \geq K^{mean} \\ 0, & else\end{cases}
\end{split}
\label{eq7}
\end{equation}

The obtained mask $M^{K}$ is used to decide which pixels can be augmented. For any data augmentation method $o' = \text{Aug}(o)$, we apply the following operation:

\begin{equation}
\tilde{o}=M^{K} \odot o+(1-M^{K}) \odot o^{\prime}
\label{final}
\end{equation}
We note that the output $\tilde{o}$ is only modified in the areas that have low relevance to the task. 

\begin{figure*}[t]
  \centering
  \vspace{-4pt}
  \includegraphics[width=1.0\linewidth]{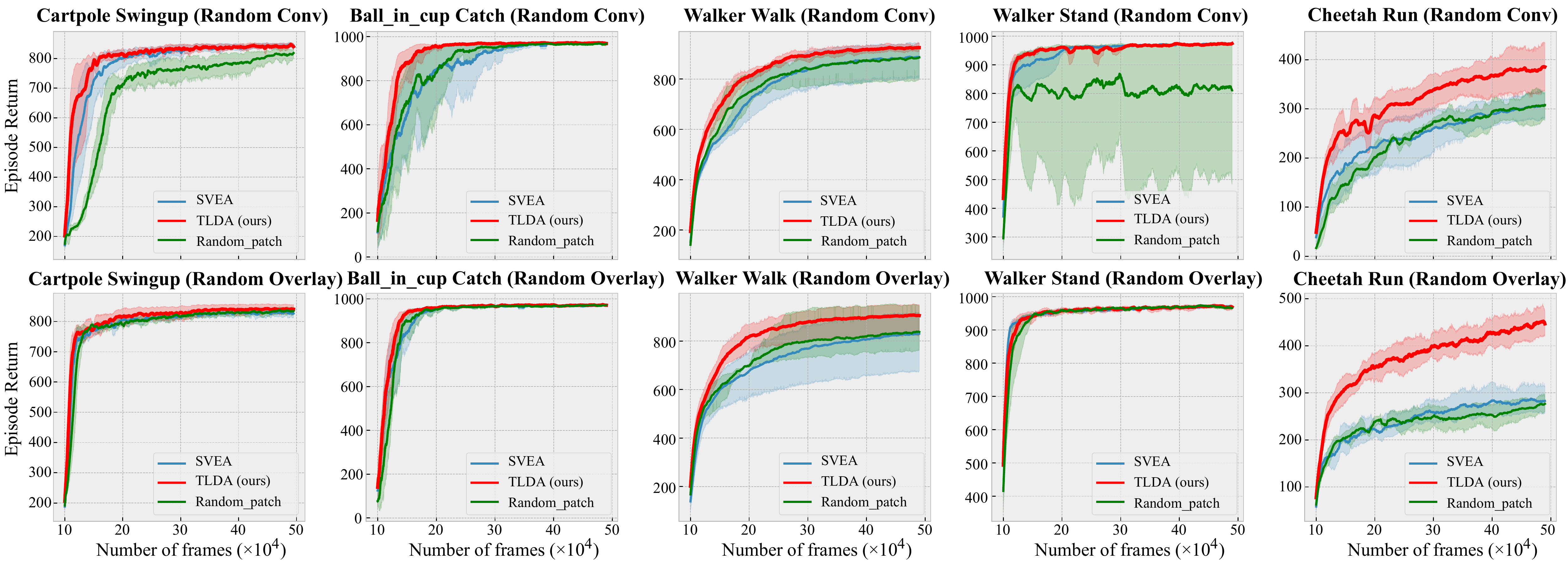}
  \vspace{-16pt}
   \caption{\textbf{Sample efficiency in training environment}. We compare TLDA, SVEA, and \textit{random patch} under two kinds of augmentations. \textit{Top row} and \textit{Bottom Row} are corresponding to \textit{Random Conv} and \textit{Random Overlay} training curves of the episode return respectively. TLDA (\textit{red line}) shows better sample efficiency on the training period. Mean and standard deviation of 5 runs.}
   \label{fig4}
  \vspace{-8pt}
\end{figure*}

\begin{algorithm}[t]
	\renewcommand{\algorithmicrequire}{\textbf{Input:}}
	\renewcommand{\algorithmicensure}{\textbf{Output:}}
	\caption{Task-aware Lipschitz Data Augmentation~(TLDA)}
	\label{alg1}
	\begin{algorithmic}[1]
	    \STATE Denote network parameters as $\theta$, $\psi$
	    \STATE Denote momentum coefficient $\tau$, Batch size $N$, Strong augmentation $\mathcal{F}$, Replay Buffer $\mathcal{B}$
        \FOR{\textit{every update iteration}}
        \STATE Sample a batch of observations $N$ from $\mathcal{B}$
        \FOR{$i=1,2, \ldots N$}
        \STATE Implement the strong augmentation: $o'_{i}=\mathcal{F}(o_{i})$
        \FOR{\textit{each pixel}}
	    \STATE Calculate the  $K_{i j}^{\pi}$ based on the distance between $\pi(\cdot \mid \Phi(o_{i},i,j))$, $\pi(\cdot \mid o_{i})$
	    \ENDFOR
	    \STATE Arrange $K_{i j}^{\pi}$ into \textit{K-matrix}
	    \STATE Get the preserved location by mask $M^{K}(o_{i})$ based on Eq (\ref{eq7})
	    \STATE Acquire the TLDA output based on Eq~(\ref{final})
	    \ENDFOR
	    \STATE Optimize $\mathcal{L}_{Q}(\theta)$ w.r.t. $\theta$
	    \STATE Update $\psi \leftarrow (1-\tau)\psi + \tau\theta$
	    \ENDFOR
	\end{algorithmic}  
	\label{algo1}
\end{algorithm}

As mentioned above, our approach tends to preserve the pixels with large $K_{ij}^{\pi}$ and augment only the pixels associated with the small ones, which adds an implicit constraint to maintain the stable output of the policy and value network. Hence, it echoes with the Optimality-Invariant State Transformation as in Definition~\ref{d1}. Figure~\ref{fig3} demonstrates the overall framework of TLDA. During the training process, the \textit{K-matrix} 
is calculated on the fly against every training step on augmented observations. Take \textit{cutout} (adding a black patch to the image) in Figure~\ref{fig3} as an example, since the corresponding \textit{K-matrix} shows that the upper part of the robot's body features large Lipschitz constants, therefore, blindly augmenting the image might touch the pixels in this area and cause catastrophic action/value changes. In contrast, TLDA preserves the critical parts of the original observations indicated by \textit{K-matrix}. It can help to maintain the stability of the action/value outputs.

\subsection{Reinforcement Learning with TLDA}
We use soft-actor-critic (SAC) as the base reinforcement learning algorithm for TLDA. Similar to previous work, we also include a regularization term $\mathcal{R}_{Q}(\theta)$ to the SAC critic loss $\mathcal{J}_{Q}\left(\theta\right)$ to handle augmented data. Our critic loss $\mathcal{L}_{Q}(\theta)$ is as follows,  where ${s}_{t}^{\text{aug}}$ is calculated by Eq (\ref{final}), and $\hat{Q}\left({s}_{t}, {a}_{t}\right)=r\left({s}_{t}, {a}_{t}\right)+\gamma \mathbb{E}_{{s}_{t+1} \sim \mathcal{P}}\left[V\left({s}_{t+1}\right)\right]$:
\vspace{10pt}
\begin{equation}
\mathcal{L}_{Q}(\theta) =\mathcal{J}_{Q}(\theta)+\lambda \mathcal{R}_{Q}(\theta)
\end{equation}

with
\vspace{-3pt}
\begin{gather*}
\begin{small}
\begin{aligned}
\mathcal{J}_{Q}(\theta) &=\mathbb{E}_{\left(s_{t}, a_{t}\right) \sim \mathcal{D}}\left[\frac{1}{2}\left(Q_{\theta}\left(s_{t}, a_{t}\right)-\hat{Q}\left(s_{t}, a_{t}\right)\right)^{2}\right] \\
\vspace{-1pt}
\mathcal{R}_{Q}(\theta) &=\mathbb{E}_{\left(s_{t}, a_{t}\right) \sim \mathcal{D}}\left[\frac{1}{2}\left(Q_{\theta}\left(s_{t}^{\text {aug }}, a_{t}\right)-\hat{Q}\left(s_{t}, a_{t}\right)\right)^{2}\right]
\end{aligned}
\end{small}
\end{gather*}

The instantiated RL algorithm are in Algorithm~\ref{algo1}, and more implementation details are summarized in Appendix~\ref{appendH}.

\section{Experiment}
In this section, we explore how TLDA can affect the agent's sample efficiency and generalization performance. We compare our method with other baselines on a wide spectrum of tasks including DeepMind control suite, CARLA simulator, as well as DeepMind Manipulation tasks. We also ablate TLDA and investigate its effect on action distributions and value estimation.  

\newcommand{\addDmcFig}[1]{\includegraphics[width=0.13\linewidth]{#1.png}}
\begin{center}
\setlength\tabcolsep{4pt}
\begin{table*} \caption{\textbf{DMC-GB Generalization Performance.} We report the episode return in test environments. The agents are trained on a fixed environment and evaluated on two unseen test environments, i.e., \textit{random colors} (\textit{Bottom}) and \textit{video backgrounds} (\textit{Top}). Our method achieves competitive or better performance in \textbf{7} out of \textbf{10} tasks.}
\label{tDMC-GB}
\centering
\begin{small}
\begin{tabular}{cccccccccc}
\cline{2-10}
\toprule[0.5mm]
Setting & \begin{tabular}[c]{@{}c@{}}DMControl\\\end{tabular} & DrQ                                        & PAD                                      & \begin{tabular}[c]{@{}c@{}}SVEA\\ (conv)\end{tabular} & \begin{tabular}[c]{@{}c@{}}SVEA\\ (overlay)\end{tabular} & \begin{tabular}[c]{@{}c@{}}SODA\\ (conv)\end{tabular} & \begin{tabular}[c]{@{}c@{}}SODA\\ (overlay)\end{tabular} & \begin{tabular}[c]{@{}c@{}}TLDA\\ (conv)\end{tabular} & \begin{tabular}[c]{@{}c@{}}TLDA\\ (overlay)\end{tabular} \\ \hline
\multirow{5}{*}{{\addDmcFig{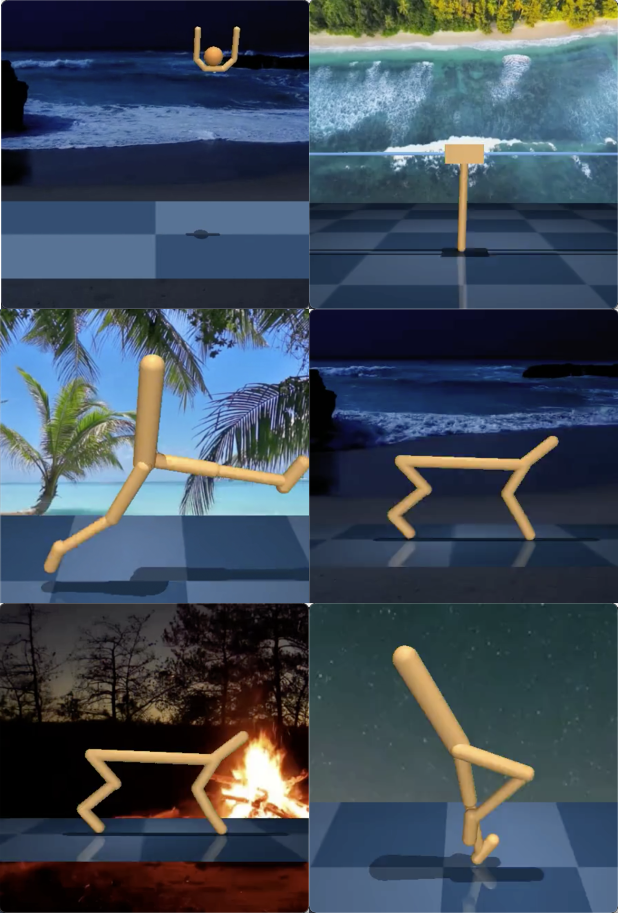}}}& \begin{tabular}[c]{@{}c@{}}Cartpole,\\ Swingup\end{tabular}          & $485$\scriptsize$\pm 105$ & $521$\scriptsize$\pm 76$ & $606$\scriptsize$\pm 85$              & \bm{$782$}\scriptsize$\pm \bm{27}$                 & $474$\scriptsize$\pm 143$             & $758$\scriptsize$\pm 62$                 & $607$\scriptsize$\pm 74$              & $671$\scriptsize$\pm 57$                                                         \\
& \begin{tabular}[c]{@{}c@{}}Walker,\\ Stand\end{tabular}              & $873$\scriptsize$\pm 83$  & $935$\scriptsize$\pm 20$ & $795$\scriptsize$\pm 70$              & $961$\scriptsize$\pm 8$                  & $903$\scriptsize$\pm 56$              & $955$\scriptsize$\pm 13$                 & ${962}$\scriptsize${\pm 15}$               & $\bm{973}$\scriptsize$\pm \bm{6}$                  \\
& \begin{tabular}[c]{@{}c@{}}Walker.\\ Walk\end{tabular}               & $682$\scriptsize$\pm 89$  & $717$\scriptsize$\pm 79$ & $612$\scriptsize$\pm 144$             & $819$\scriptsize$\pm 71$                 & $635$\scriptsize$\pm 48$              & $768$\scriptsize$\pm 38$                 & \bm{$873$}\scriptsize$\pm \bm{34}$              & ${868}$\scriptsize$\pm {63}$                                                         \\
& \begin{tabular}[c]{@{}c@{}}Ball\_in\_cup,\\ Catch\end{tabular}       & $318$\scriptsize$\pm 157$  & $436$\scriptsize$\pm 55$ & $659$\scriptsize$\pm 110$             & $871$\scriptsize$\pm 106$                & $539$\scriptsize$\pm 111$             & ${875}$\scriptsize$\pm {56}$                 & $\bm{887}$\scriptsize${\pm \bm{58}}$              &  ${855}$\scriptsize${\pm 56}$                                                        \\
& \begin{tabular}[c]{@{}c@{}}Cheetah,\\ Run\end{tabular}               & $102$\scriptsize$\pm 30$  & $206$\scriptsize$\pm 34$ & $292$\scriptsize$\pm 32$              & $249$\scriptsize$\pm 20$                 & $229$\scriptsize$\pm 29$              & $223$\scriptsize$\pm 32$                 & $\bm{356}$\scriptsize${\pm \bm{52}}$              &  ${336}$\scriptsize${\pm 57}$                                                        \\ \midrule[0.3mm]

\multirow{5}{*}{{\addDmcFig{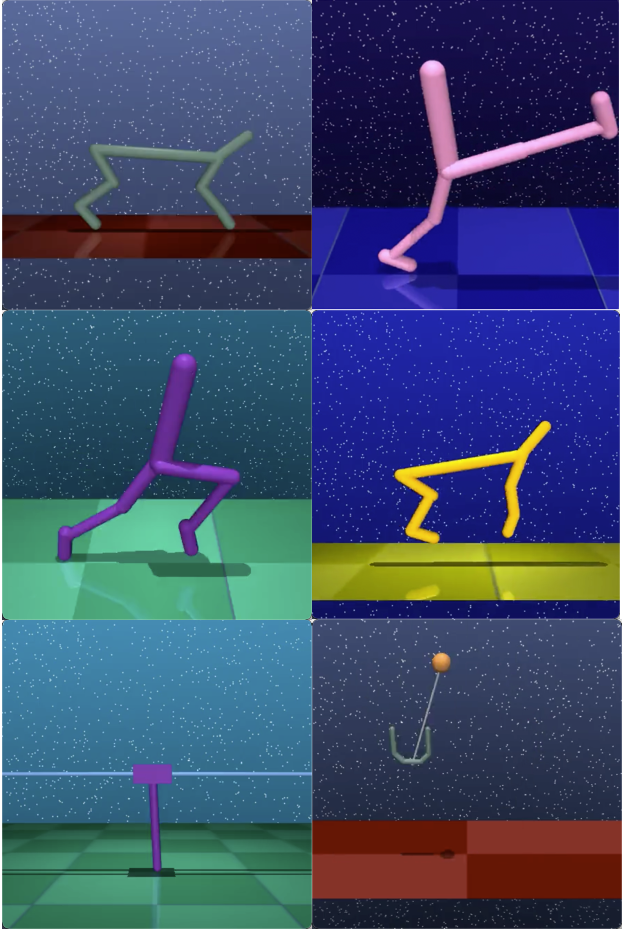}}}& \begin{tabular}[c]{@{}c@{}}Cartpole,\\ Swingup\end{tabular}         & $586$\scriptsize$\pm 52$  & $630$\scriptsize$\pm 63$ & $\bm{837}$\scriptsize$\pm \bm{23}$              & $832$\scriptsize$\pm 23$                 & $831$\scriptsize$\pm 21$              & $805$\scriptsize$\pm 28$                 & $748$\scriptsize$\pm 40$              & ${760}$\scriptsize$\pm{60}$                 \\
& \begin{tabular}[c]{@{}c@{}}Walker,\\ Stand\end{tabular}             & $770$\scriptsize$\pm 71$  & $797$\scriptsize$\pm 46$ & ${942}$\scriptsize$\pm {26}$              & $933$\scriptsize$\pm 24$                 & $930$\scriptsize$\pm 12$              & $893$\scriptsize$\pm 12$                 & $919$\scriptsize$\pm 24$              & $\bm{947}$\scriptsize$\pm \bm{26}$                 \\
& \begin{tabular}[c]{@{}c@{}}Walker.\\ Walk\end{tabular}              & $520$\scriptsize$\pm 91$  & $468$\scriptsize$\pm 47$ & $760$\scriptsize$\pm 145$             & $749$\scriptsize$\pm 61$                 & $697$\scriptsize$\pm 66$              & $692$\scriptsize$\pm 68$                 & $753$\scriptsize$\pm 83$             & $\bm{823}$\scriptsize$\pm \bm{58}$                 \\
& \begin{tabular}[c]{@{}c@{}}Ball\_in\_cup,\\ Catch\end{tabular}      & $365$\scriptsize$\pm 210$ & $563$\scriptsize$\pm 50$ & $\bm{961}$\scriptsize$\pm \bm{7}$               & $959$\scriptsize$\pm 5$                  & $892$\scriptsize$\pm 37$              & $949$\scriptsize$\pm 19$                 & $932$\scriptsize$\pm 32$              & ${930}$\scriptsize$\pm {40}$                 \\
& \begin{tabular}[c]{@{}c@{}}Cheetah,\\ Run\end{tabular}              & $100$\scriptsize$\pm 27$  & $159$\scriptsize$\pm 28$ & $264$\scriptsize$\pm 51$              & $273$\scriptsize$\pm 23$                 & $294$\scriptsize$\pm 34$              & $238$\scriptsize$\pm 28$                 & $\bm{371}$\scriptsize$\pm \bm{51}$              & ${358}$\scriptsize$\pm {25}$                  \\ \hline
\end{tabular}
\end{small}
\end{table*}
\end{center}

\vspace{-8pt}
\subsection{Evaluation on DeepMind Control Suite}

\quad \textbf{Setup.} We implement our method with SAC as the base algorithm. Convolution Neural Networks are used for the image inputs. We include a detailed description of all hyper-parameters and the architecture in Appendix~\ref{appendH}. For comparison, we mainly consider two augmentation ways applied in the prior state-of-the-art methods: \textit{random convolution} (passing input through a random convolutional layer) and \textit{random overlay} (linearly combining the observation $o$ with the extra image $\mathcal{I}$, $\phi(o) = \alpha o + ( 1 - \alpha ) \mathcal{I})$.

\textbf{Baselines.} We benchmark TLDA against the following state-of-the-art methods: (1) \textbf{DrQ}~\cite{kostrikov2020image}: SAC with weak augmentation (\textit{random shift}); (2) \textbf{PAD}~\cite{hansen2020self}: adding an auxiliary task for adapting to the unseen environment; (3) \textbf{SODA}~\cite{hansen2021generalization}:  maximizing the mutual information between latent representation  by employing a BYOL-like~\cite{grill2020bootstrap} architecture;~(4) \textbf{SVEA}~\cite{hansen2021stabilizing}: modifying the form of Q-target. We run 5 random seeds and report the mean and standard deviation of episode rewards.

\textbf{Sample efficiency under strong augmentations.} We compare the sample efficiency with SVEA to exhibit the effectiveness of TLDA. 

We also include another baseline that preserves random patches from the un-augmented observation as opposed to TLDA that preserves task-related parts. We call this baseline \textit{random patch}.
By contrast, SVEA only uses the strong augmentation method but retains no raw pixel. Figure~\ref{fig4} demonstrates that TLDA achieves better or comparable asymptotic performance in the training environment than baselines on DM-control suite while having better sample efficiency. The results also indicate that \textit{random patch} will hinder the performance in some tasks. We reckon that since \textit{random patch} does not have any pixel-to-task relevance knowledge, it inevitably destroys the image's integrity and even leads to further distortion to the observations after data augmentation. Therefore, blindly keeping the original observation's information cannot improve the agent's training performance. It is the retention of areas with larger Lipschitz constants, instead of random original areas, that boosts the sample efficiency of training agents.

\textbf{Generalization Performance.} We evaluate the agent's generalization ability on two settings from DMControl-GB~\cite{hansen2021generalization}:~(i)~random colors of the background and agent;~(ii)~dynamic video backgrounds. Results are shown in Table~\ref{tDMC-GB}. TLDA outperforms prior state-of-the-art methods in \textbf{7} out of \textbf{10} instances. The agent trained with TLDA is able to acquire a good robust policy in different unseen environments. Meanwhile, we notice that prior methods are sensitive to different kinds of augmentations, which makes their testing performance varies dramatically. On the contrary, our method with task-aware observations is more stable and not susceptible to this issue. 

\begin{figure}[t]
  \centering
  \includegraphics[width=1.0\linewidth]{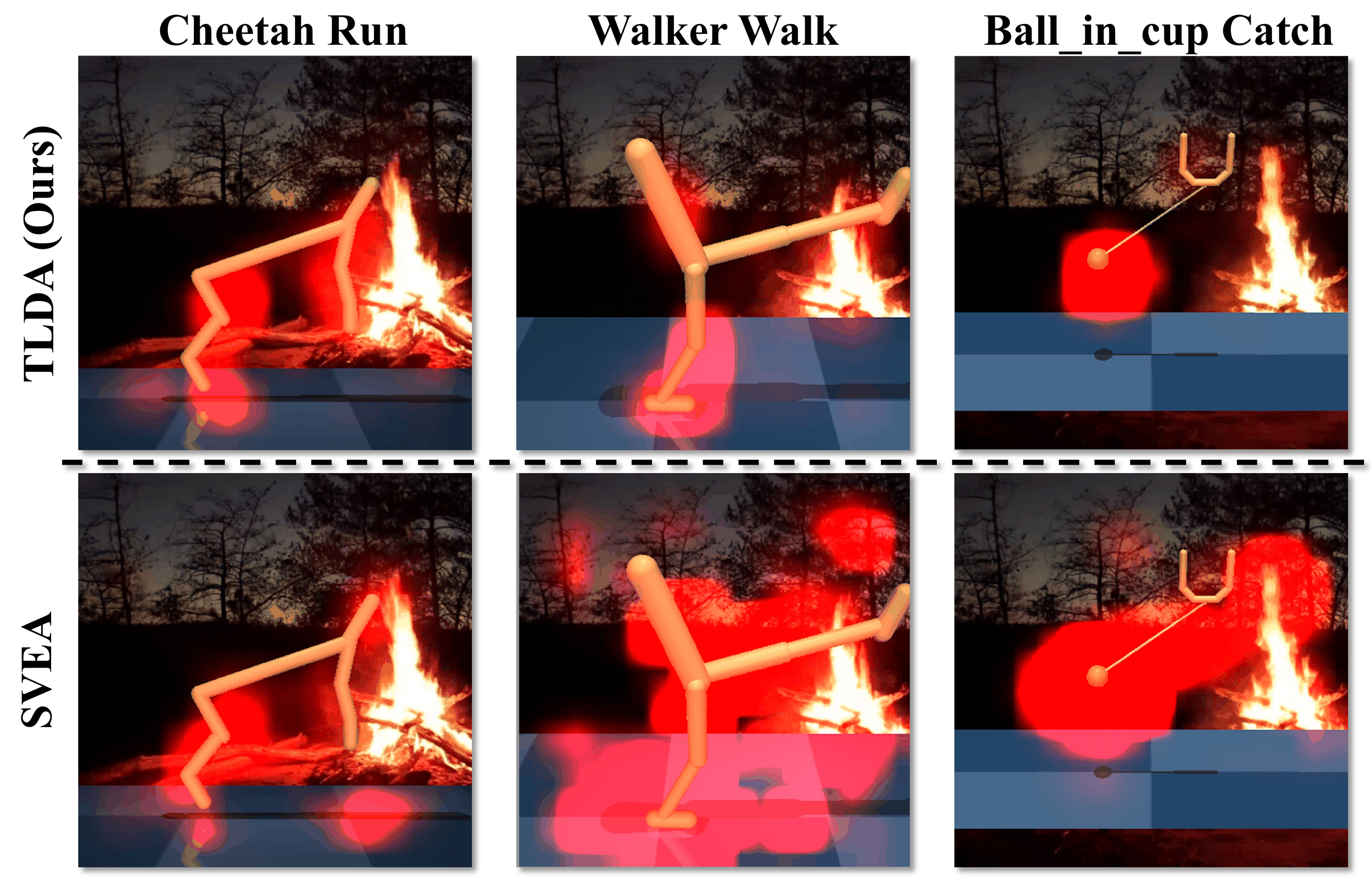}
  \vspace{-8pt}
   \caption{\textbf{Visualization of \textit{K-matrix} in Generalization}. We visualize the same observation frame of \textit{K-matrix} (red color) about SVEA (\textit{Bottom Row}) and TLDA (\textit{Top Row}) during generalization . It exhibits that the \textit{K-matrix} calculated by SVEA will highlight the video background while TLDA still focuses on the robot body.}
   \label{fig8}
  \vspace{-15pt}
\end{figure}

\textbf{Qualitative Results of TLDA.} As shown in Figure~\ref{fig8}, from the \textit{K-matrix} on the test environments, agents trained by TLDA will give larger Lipschitz constants on the robot's body while the SVEA agents are prone to focus on the lighting visual background. Our method is capable of learning the main factors that influence the performance and neglecting the irrelevant areas that hinder generalization.

\textbf{Effect on Action Distribution and Q-estimation.} In this section, we analyze how TLDA influences the output of the policy and value networks. Given a DrQ agent trained in the original environment, we assess the Q-value estimation and the action distribution under different augmentation. To get a better understanding of this issue, we visualize the action distribution of the agent under different augmentation methods, as shown in Figure~\ref{fig1}. For weak augmentation, although its action distribution is closest to the un-augmented one (Figure~\ref{fig:action-error-weak}), it cannot improve generalization, as shown in Table~\ref{tDMC-GB}(DrQ). Strong augmentation, on the other hand, will cause an obvious distribution shift (Figure~\ref{fig:action-error-strong}), thus significantly hindering the training process.

We find that TLDA has a closer action distribution than simply applying strong augmentation (Figure~\ref{fig:action-error-ours}) by using the Lipschitz constant to identify and preserve the task-aware areas. Furthermore, as shown in Figure~\ref{fig2}, we find that the Q-estimation of TLDA has a lower variance than that of naively applying strong augmentation. These two results illustrate that TLDA has the potential to achieve higher sample efficiency in training and learn a more robust policy to perform well in unseen environments.

\subsection{Evaluation on Autonomous Driving in CARLA}
\begin{figure}[h]
  \centering
  \vspace{-15pt}
  \includegraphics[width=0.75\linewidth]{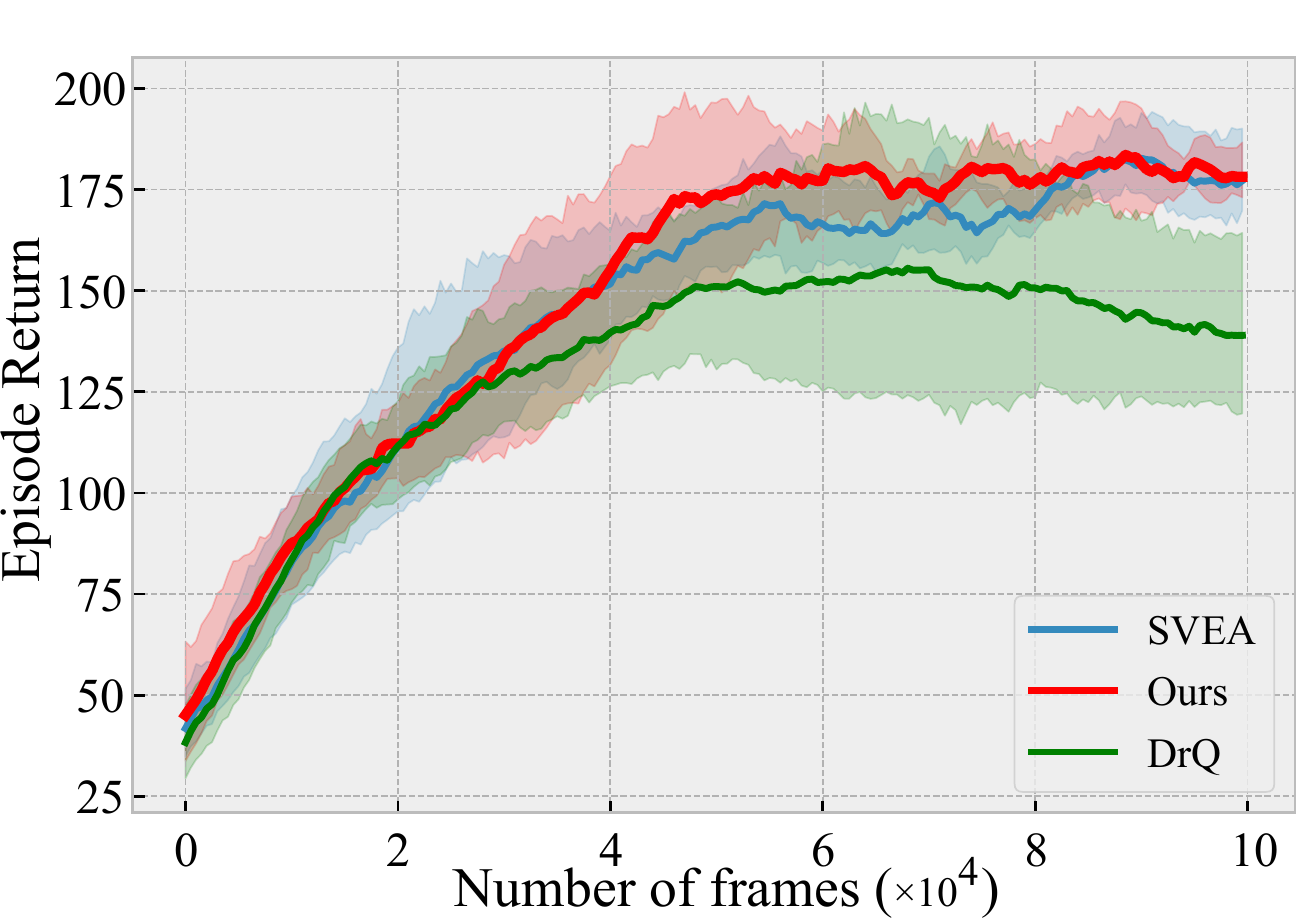}
   \caption{\textbf{CARLA Training Performance}. We evaluate three algorithms across 5 seeds. TLDA (\textit{red line}) achieves better performance than SVEA (\textit{blue line}) and DrQ (\textit{green line}) in sample efficiency.}
   \label{fig:carla}
   \vspace{-13pt}
\end{figure}

To further evaluate the TLDA's performance, we apply this method in the tasks with more realistic observations: autonomous driving in the CARLA simulator. In our experiment, we use one camera as our input observation for driving tasks, where the goal of the agent is to drive along a curvy road as far as possible in 1000 time-steps without colliding with the moving vehicles, pedestrians and barriers. We adapt the reward function and train an agent under the weather with the same setting from previous work~\cite{zhang2020learning}. The training results are shown in Figure~\ref{fig:carla}. We find that our method achieves the best training sample efficiency. For generalization, CARLA provides different weather conditions with built-in parameters. We evaluate our method in 4 kinds of weather with different lighting conditions, realistic raining and slipperiness. Results are in Table~\ref{carlat}, where we choose the success rate for reach 100m distance as the driving evaluation metric. TLDA outperforms all base algorithms in both sample efficiency and generalization ability with a more stable driving policy. Additional results are in Appendix~\ref{carla}.

\vspace{-2pt}
\begin{table}[h] 
\caption{\textbf{CARLA Driving}. We report the success rate for reaching 100m distance under the unseen weather during 250 episodes across 5 seeds for each weather. (50 episodes for each seed.)}
\vspace{-5pt}
\centering
\begin{small}
\begin{tabular}{cccc}
\toprule[0.5mm]
Setting         & DrQ   & SVEA  & Ours  \\ \midrule[0.3mm]
Training        & $24\%$  & $49\%$ & $\bm{52\%}$ \\
Wet Noon        & $0.8\%$  & $8.8\%$  & \bm{$18\%$}  \\
SoftRain noon  & {$0.4\%$} & {$1.2\%$}   & \bm{$7.6\%$}  \\
Wet Sunset  & $0.8\%$  & {$1.6\%$}   & \bm{$9.2\%$}  \\
MidRain Sunset    & $0.0\%$      & {$5.2\%$}  & \bm{$12\%$}  \\ \hline
\end{tabular}
\end{small}
\label{carlat}
\vspace{-5pt}
\end{table}

\vspace{-5pt}
\subsection{Evaluation on DMC Manipulation Tasks}
Robot  manipulation  is  another set of challenging  and  meaningful  tasks  for  visual  RL. DM control~\cite{tunyasuvunakool2020dm_control} provides a set of configurable manipulation tasks with a robotic Jaco arm and snap-together bricks. We consider two tasks for experiments: \textit{reach} and \textit{push}. More details are in Appendix~\ref{mani}.

All the agents are trained on the default background and evaluated on different colors of arms and platforms. Training results and generalization performance are shown in Appendix~\ref{mani:result} and Table~\ref{tmani}. The results show that our method can be adapted to the unseen environments more appropriately. The \textit{Modified Platform} and \textit{Modified Both} setting are challenging for agents to discern the target objects from the noisy backgrounds. SVEA under strong data augmentation suffers from instability and divergence for training, while TLDA can augment the pixel in a task-aware manner, thus further maintaining the training stability. Despite that DrQ shows better training performance, it barely generalizes to the environments with different visual layouts. In summary, sample efficiency and generalization performance contribute to exhibiting the superiority of the proposed algorithm.
\begin{table}[h] 
\caption{ \textbf{DMC Manipulation Tasks}. We evaluate the episode return in different modified (M) visual setting. M in the \textit{Setting} column means: \textbf{\textit{Modified}}. TLDA can better focus on the aim objects in the noisy and colorful visual backgrounds.}
\vspace{-5pt}
\centering
\begin{small}
\begin{tabular}{ccccc}
\toprule[0.5mm]
Task                   & Setting    & DrQ   & SVEA & Ours  \\ \midrule[0.3mm]
\multirow{4}{*}{Reach} & Training   & $\bm{136}$ \scriptsize$\pm \bm{20}$ & $49$ \scriptsize$\pm 48$  & $124$ \scriptsize$\pm 32$\\
                       & M Arm      & $\bm{68}$ \scriptsize$\pm \bm{20}$ & $21$ \scriptsize$\pm 25$  & $55$ \scriptsize$\pm 21$ \\
                       & M Platform & $0.8$ \scriptsize$\pm 1.3$  & $24$ \scriptsize$\pm 25$ & $\bm{89}$ \scriptsize$\pm \bm{40}$  \\
                       & M Both     & $1$ \scriptsize$\pm 2$  & $13$ \scriptsize$\pm 14$ & $\bm{36}$ \scriptsize$\pm \bm{25}$  \\ \midrule
\multirow{4}{*}{Push}  & Training   & $\bm{141}$ \scriptsize$\pm \bm{47}$      & ${42}$ \scriptsize$\pm {40}$     & ${109}$ \scriptsize$\pm {27}$      \\
                       & M Arm      & $\bm{88}$ \scriptsize$\pm \bm{52}$      &  ${21}$ \scriptsize$\pm {16}$    &  ${60}$ \scriptsize$\pm {43}$     \\
                       & M Platform &  ${4}$ \scriptsize$\pm {1}$     & ${34}$ \scriptsize$\pm {28}$     & $\bm{95}$ \scriptsize$\pm \bm{33}$      \\
                       & M Both     &  ${5}$ \scriptsize$\pm {1}$     &  ${32}$ \scriptsize$\pm {20}$    &  $\bm{56}$ \scriptsize$\pm \bm{42}$     \\ \hline
\end{tabular}
\end{small}
\label{tmani}
\vspace{-10pt}
\end{table}

\section{Conclusion}
In this paper, we propose \textbf{T}ask-aware \textbf{L}ipschitz \textbf{D}ata \textbf{A}ugmentation (TLDA) for visual RL, which can reliably identify and augment pixels that are not strongly correlated with the learning task while keeping task-related pixels untouched. This technique aims to provide a principled mechanism for boosting the generalization ability of RL agents and can be seamlessly incorporated into various existing visual RL frameworks. Experimental results on three challenging benchmarks confirm that, compared with the baselines, TLDA not only features higher sample efficiency but also helps the agents generalize well to the unseen environments.

\bibliographystyle{arxiv_natbib}
\bibliography{anthology}

\clearpage

\onecolumn
\appendix
\section{The Proof of Proposition 1} \label{a1}
\textbf{Proposition 1}\emph{ We consider an MDP $\mathcal{M}$ and an augmentation method $\phi$. Let $\pi$ be a policy over $\phi$. Suppose the rewards are bounded by $r_{max}$ such that $\forall a \in \mathcal{A}, \forall s \in \mathcal{S}, |r(s, a)| \leq r_{max}.$ Then for $\phi$ and $\pi$, the following inequality holds :
\begin{equation}
\begin{aligned}
\left| Q ^ { \pi } ( s , a ) - Q ^ { \pi } ( \phi ( s ) , a ) \right| \leq 2r_{max}\frac{(K_{\pi}\left\| d(\phi) \right\| _ { \infty }+ 1)}{1-\gamma}
\end{aligned}
\end{equation}
}

\emph{Proof.} We are interested in the Q-value function which is defined as follows:

\begin{equation}
\begin{aligned} Q ^ { \pi } ( s , a ) & = E _ { \pi } \left[ r _ { t + 1 } + \gamma r _ { t + 2 } + \cdots \mid A _ { t } = a , S _ { t } = s \right] \\ & = \sum _ { s } \sum _ { a } \sum _ { t } \gamma ^{t} p ^ { t } ( s ) \pi ( a | s ) r _ { t } ( s , a ) \end{aligned}
\end{equation}

Let $p_{\phi}^{t}(s)$ be the probability of visiting state $\phi(s)$ at time $t$, $\pi_{\phi}(a|s)$ be the probability of doing action $a$ in state $\phi(s)$, thus

\begin{equation}
Q ^ { \pi } ( \phi ( s ) , a ) = \sum _ { s } \sum _ { a } \sum _ { t } \gamma ^{t} p _ { \phi } ^ { t } ( s ) \pi _ { \phi } ( a|s ) r _ { t } ( s , a )
\end{equation}

By these definitions, we can write that:
\begin{equation}
\begin{aligned}
\left| Q ^ { \pi } ( s , a ) - Q ^ { \pi } ( \phi ( s ) , a ) \right| 
&\leq r _ { max } \sum _ { s } \sum _ { a } \sum _ { t } \gamma ^ { t }  |p _ { \phi } ^ { t } ( s ) \pi _ { \phi } \left( a | s ) - p ^ { t } ( s ) \pi ( a|s ) \mid \right. \\
& = r _ { max } \sum _ { s } \sum _ { a } \sum _ { t } \gamma ^ { t }  |p _ { \phi } ^ { t } ( s ) \pi _ { \phi } ( a | s ) +p _ { \phi } ^ { t } ( s )\pi ( a|s )-p _ { \phi } ^ { t } ( s )\pi ( a|s )- p ^ { t } ( s ) \pi ( a|s )| \\
& \leq r _ { max } \sum _ { s } \sum _ { a } \sum _ { t } \gamma^{t}(p_{\phi}^{t}(s)|\pi_{\phi}(a|s)-\pi(a|s)|+\pi(a|s)|p_{\phi}^{t}(s)-p^{t}(s)|) \\
& = r _ { max }(\sum _ { s } \sum _ { a }\sum _ { t } \gamma^{t}(p_{\phi}^{t}(s)|\pi_{\phi}(a|s)-\pi(a|s)|)+ 2\sum _ { t }\gamma^{t}D_{TV}(p_{\phi}^{t}(\cdot)|p^{t}(\cdot))) )\\
& \leq 2r_{max}\sum_{t}\gamma^{t}(\max_{s}D_{TV}(\pi_{\phi}(\cdot|s)||\pi(\cdot|s))+D_{TV}(p_{\phi}^{t}(\cdot)||p^{t}(\cdot)))
\end{aligned}
\label{eq20}
\end{equation}

$D_{TV}(P||Q) = \frac{1}{2} \sum_{a \in \mathcal{A}}|P(a)-Q(a)|$. This above inequality shows that the Q-value estimation is bounded by two total variation distance (the smaller the total variation distance is, the closer the Q-value estimation will be).

\vspace{5pt}
\textbf{Lemma2} \emph{For a given state $s$, and a given policy $\pi(\cdot |s)$ under the data augmentation method $\phi$, we assume the state space is equipped with a distance metric $d(\cdot, \cdot)$, the following bound holds:}

\begin{equation}
\max_{s}D_{TV}(\pi_{\phi}(\cdot |s)||\pi(\cdot|s))\leq K_{\pi}\left\| d(\phi) \right\| _ { \infty }
\end{equation}

\emph{Proof.}

\begin{equation}
\begin{aligned}
&\max_{s}D_{TV}(\pi_{\phi}(\cdot |s)||\pi(\cdot|s))\\
& = \max_{s}\frac{D_{TV}(\pi_{\phi}(\cdot |s)||\pi(\cdot|s))}{d(\phi(s),s)}d(\phi(s),s)\\
& \leq \max_{s}\frac{D_{TV}(\pi_{\phi}(\cdot |s)||\pi(\cdot|s))}{d(\phi(s),s)} \max_{s}d(\phi(s),s) \\
& = K_{\pi}\left\| d(\phi) \right\| _ { \infty }
\end{aligned}
\end{equation}

We also note that the $D_{TV}(\cdot | \cdot) \in [0,1]$, so according to the bound of Eq (\ref{eq20}) we have that

\begin{equation}
\begin{aligned}
\left| Q ^ { \pi } ( s , a ) - Q ^ { \pi } ( \phi ( s ) , a ) \right| &\leq 2r_{max}\sum_{t}\gamma^{t}(\max_{s}D_{TV}(\pi_{\phi}(\cdot|s)||\pi(\cdot|s))+D_{TV}(p_{\phi}^{t}(\cdot)||p^{t}(\cdot)))\\
&\leq 2r_{max}\sum_{t}\gamma^{t}(K_{\pi}\left\| d(\phi) \right\| _ { \infty } + 1) \\
& \leq 2r_{max}\frac{(K_{\pi}\left\| d(\phi) \right\| _ { \infty } + 1)}{1-\gamma}
\end{aligned}
\end{equation}

\section{Implementation Details} \label{appendH}
In this section, we provide details of our algorithm's implementation and hyperparameter setting. Table~\ref{hyper} exhibits the hyper-parameters of TLDA in three benchmarks. It should be noticed that if we calculate the Lipschitz constant for each pixel, it will increase the computation complexity and take more time for training. Therefore, we calculate the Lipschitz constant every 5 pixels during training for one observation, which is enough to  acquire good performance. We choose Gaussian blur as the kernel and implement the 2D Gaussian centered at $(i, j)$ as Mask $M(i, j)$. Since we only change a few chosen perturbed pixels (near the pixel$(i,j)$ by $M(i, j)$) in the whole image for every location $(i, j)$ when calculating the distance between $\Phi(o,i,j)$ and $o$, we can approximate that this metric is not relevant to the specific location $(i,j)$; thus, the Lipschitz constant can be proportional to the distance between two action distributions: $\pi(\cdot \mid \Phi(o,i,j))$, $\pi(\cdot \mid o)$.

\begin{figure}[ht]
  \centering
  \includegraphics[width=0.7\linewidth]{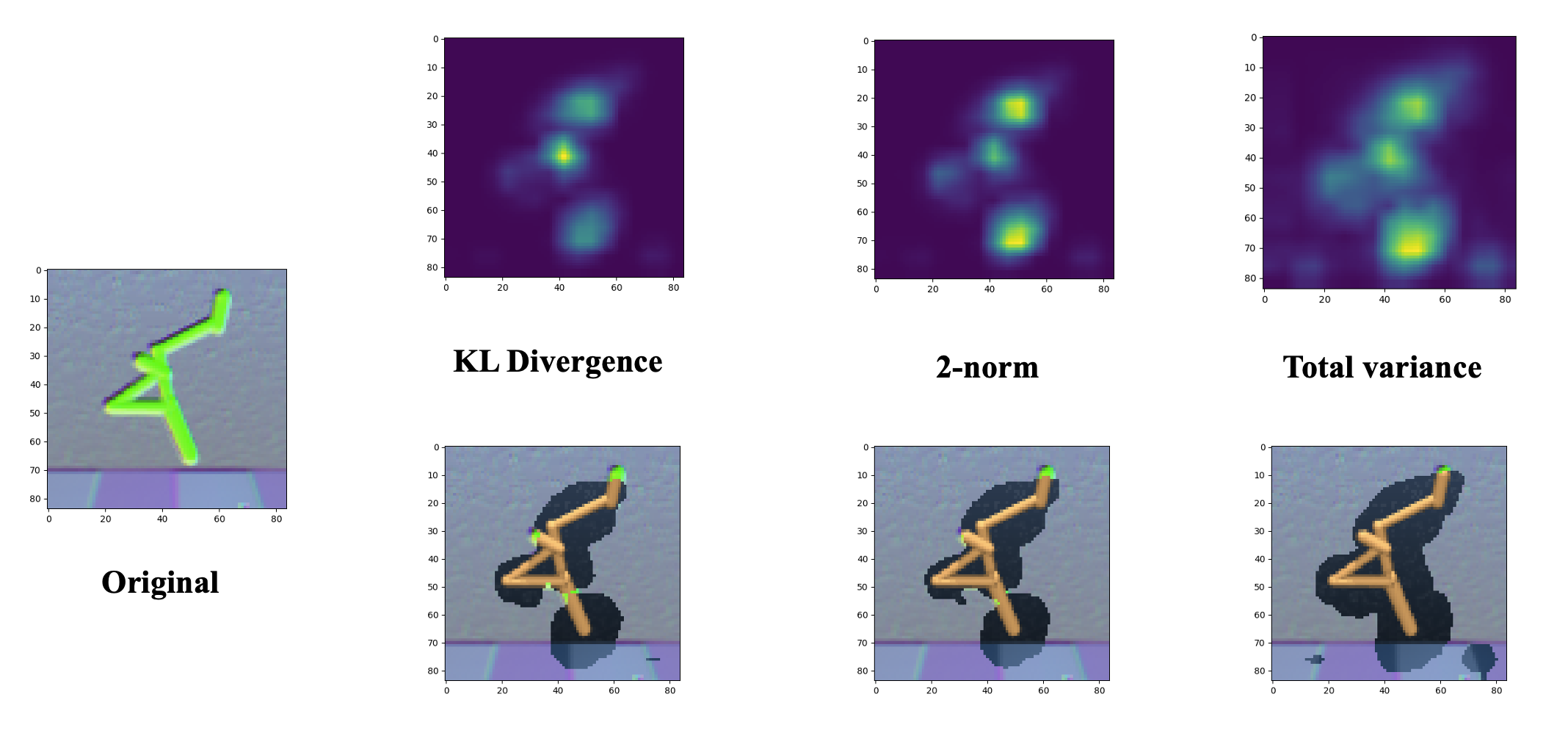}
  \vspace{-10pt}
   \caption{\textbf{{Different metrics for TLDA}}. We compare the \textit{K-matrix} under different metrics. This figure demonstrates that the $\ell_{2}$ distance is sharper and more concentrated while the total-variance distance has the Blurry effect as well as  KL divergence is darker.}
   \label{fignorm}
\end{figure}

Meanwhile, we compare different metrics for TLDA, as shown in Figure~\ref{fignorm}. The result shows that $\ell_{2}$ distance is sharper and more concentrated while the total-variance distance has the Blurry effect, as well as  KL divergence is darker. So empirically, we choose $\ell_{2}$ distance as our metric to calculate the Lipschitz constant during all experiments. DrQ shows that \textit{shifting} is an effective method to improve sample efficiency. Therefore we apply \textit{shifting} to the observation at first. The linear combination factor $\alpha$ of \textit{random overlay} is $0.5$ with the dataset in DMC-GB. We summarize our method in Algorithm \ref{algo1}. The inner two \textit{for loops} can be computed in parallel.

\begin{table}[ht]

\centering
\caption{Hyperparameter about TLDA in 3 benchmarks.}
\begin{tabular}{cccc}
\toprule[0.5mm]
Hyperparameter                             & DMControl-GB                                                                 & CARLA                     & Manipulation Tasks        \\ \hline
Input dimension                            & 9 $\times$ 84 $\times$ 84                                                    & 9 $\times$ 84 $\times$ 84 & 9 $\times$ 84 $\times$ 84 \\
Stacked Frames                             & 3                                                                            & 3                         & 3                         \\
Discount factor $\gamma$                   & 0.99                                                                         & 0.99                      & 0.99                      \\
Action repeat                              & 8(cartpole) 4(otherwise)                                                     & 4                         & 2                         \\
Actor learning rate                        & \begin{tabular}[c]{@{}c@{}}5e-4(walker walk) \\ 1e-3(otherwise)\end{tabular} & 1e-3                      & 3e-5                      \\
Critic learning rate                       & \begin{tabular}[c]{@{}c@{}}5e-4(walker walk) \\ 1e-3(otherwise)\end{tabular} & 1e-3                      & 3e-5                      \\
Random cropping padding                    & 4                                                                            & 4                         & 4                         \\
Batch size                                 & 128                                                                          & 128                       & 128                       \\
Regularization term $\lambda$ & 1                                                                            & 1                         & 1                         \\
Training step                              & 500k                                                                         & 100k                      & 500k                      \\
Replay buffer size                         & 500,000                                                                      & 100,000                   & 500,000                   \\
Encoder conv layers                        & 4                                                                            & 4                         & 4                         \\
Optimizer($\theta$)                        & Adam                                                                         & Adam                      & Adam                      \\ \hline
\end{tabular}
\label{hyper}
\end{table}

\section{Environment Details}
\begin{figure}[ht]
  \centering
  \includegraphics[width=0.5\linewidth]{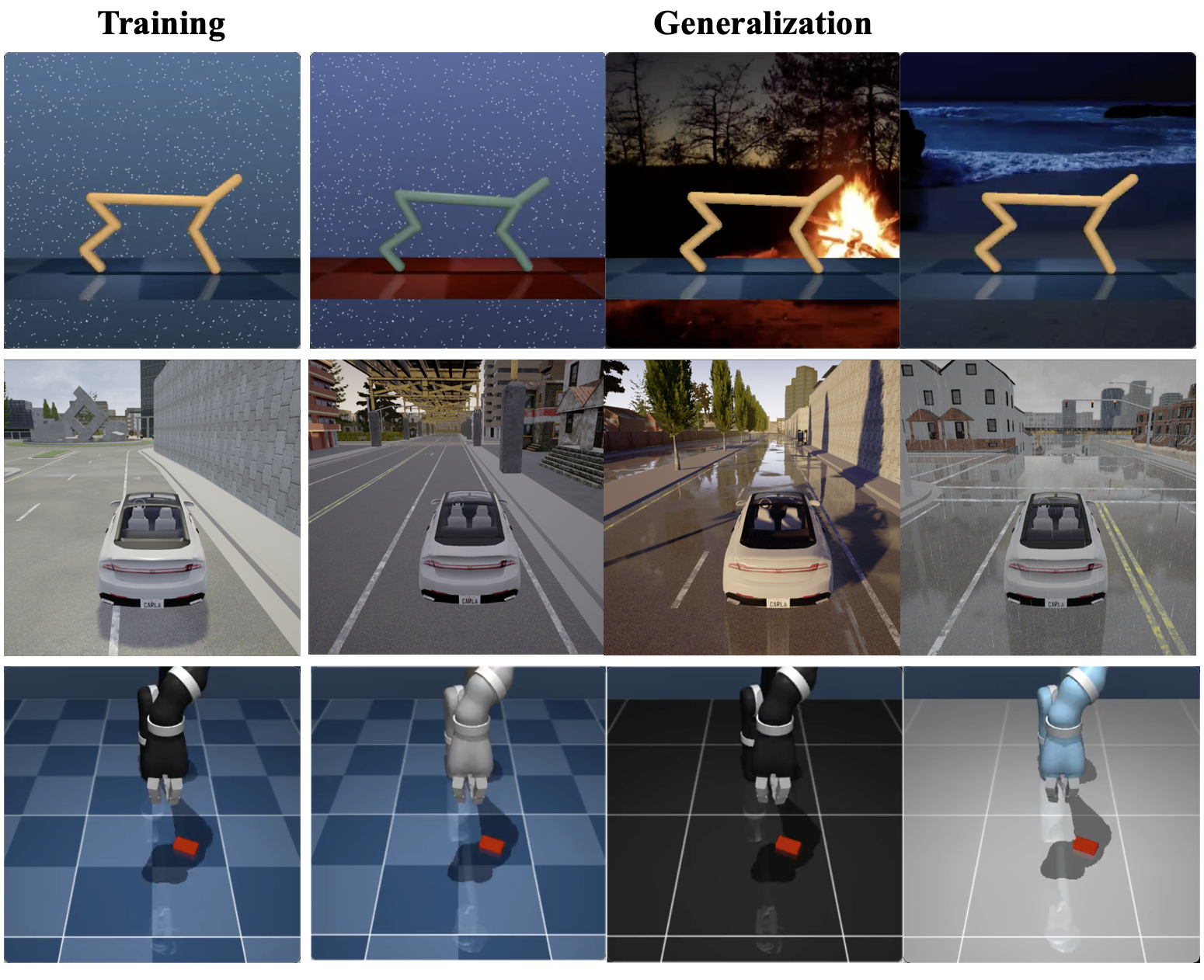}
   \caption{\textbf{Three Benchmarks for visualization.} \textit{Top} to \textit{Bottom}: DMC Control Suite, CARLA simulator autonomous driving, and DMC manipulation tasks.}
   \label{fig0}
\end{figure}

We conduct our method on three challenging visual control benchmarks, as shown in Figure~\ref{fig0}.

\subsection{DeepMind Control Suite}
DMC-GB, a popular benchmark modified from DMControl suite is introduced in ~\cite{hansen2021generalization} for visual RL. We use 5 typical tasks that support random color and dynamic video backgrounds. The detailed setting is listed in Table~\ref{hyper}.

\subsection{CARLA}
CARLA is a widely-used autonomous driving simulator. In our experiment, we choose \textit{highway CARLA Town4} as the map for the driving task. The goal of the training agent is to drive along a highway road as far as possible under diverse weather conditions. We choose a stable version of CARLA 0.9.6 \cite{dosovitskiy2017carla} and use the reward function and network architecture in \cite{zhang2020learning}. We use one camera as our input observation, which is an $84 \times 84 \times 3$ image. The action is composed of thrusting and steering these two continuous controls. We choose \textit{random overlay} as the strong augmentation method during training, whose linear combination factor $\alpha=0.5$ with the dataset from DMC-GB.

\subsection{DeepMind Manipulation Task} \label{mani}
DeepMind Manipulation task is introduced in \cite{tunyasuvunakool2020dm_control} for robot continuous controls, which provides a Kinova robotic arm and a list of objects for building reward functions. 

\begin{figure*}[t]
    \centering
    \setlength{\abovecaptionskip}{5pt}
    \subfigure[\textit{Reach} \label{fig:reach}]
        {\includegraphics[width=0.23\linewidth]{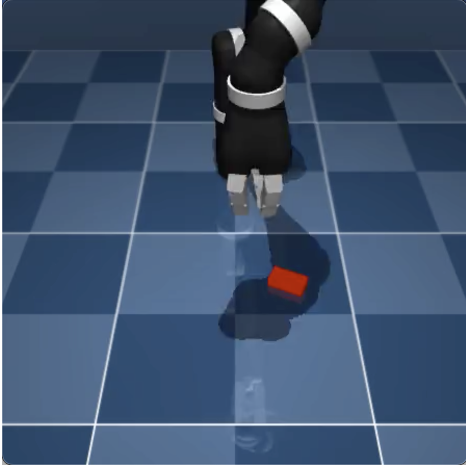}}
    \hspace{5mm}
    \subfigure[\textit{Push} \label{fig:push}]
        {\includegraphics[width=0.23\linewidth]{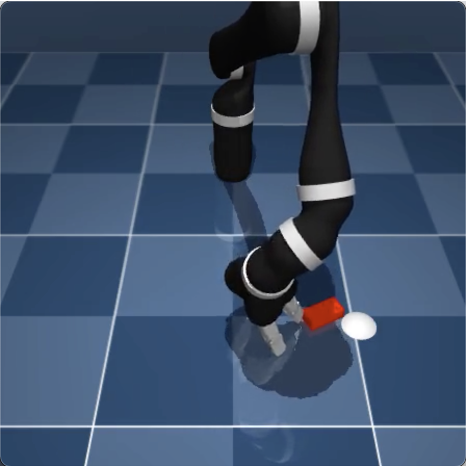}}
\vspace{-5pt}
\caption{\textbf{Two tasks of Manipulation.} (a) is the \textit{reach} task; the goal of the agent is to reach the red brick. (b) is the \textit{push} task;  we add a white point as a goal position; the goal of the agent is to push the red brick to the white position.}
\label{fig:mani_show}
\vspace{-10pt}
\end{figure*}

We additionally consider two tasks for the experiments: 
\begin{itemize}
    \item \textbf{reach}: the agent needs to reach the shown red brick by manipulating the arm;
    \item \textbf{push}: the goal is to push the red bricks to the position of a white mark point;
\end{itemize}  
The input observation is the stacked RGB images of $84 \times84$ pixels. There are two available observation versions: \textit{feature vectors} and \textit{pixel images}. All environments return a reward $r(s,a )\in [0,1]$ per step, and have an episode time limit in 10 seconds. {It is challenging for the RL algorithm based on SAC to perform well in these tasks without adding other useful tools.} Therefore, we choose the task \textit{reach\_duplo\_vision}, which aims to move the arm to a brick resting on the ground, and create a task \textit{push\_brick\_vision}, whose goal is to push a brick to a goal position. The reward function of \textit{push} is based on the \textit{reach} task, and we add a reward term about the distance between the red brick and goal position. The closer distance between them, the higher reward the agent will achieve. We visualize the observations of two tasks, as shown in  Figure~\ref{fig:mani_show}. We choose \textit{random overlay} as the strong augmentation way during training whose linear combination factor $\alpha=0.5$ with the dataset from DMC-GB.

\begin{figure}[ht]
  \centering
  \includegraphics[width=0.55\linewidth]{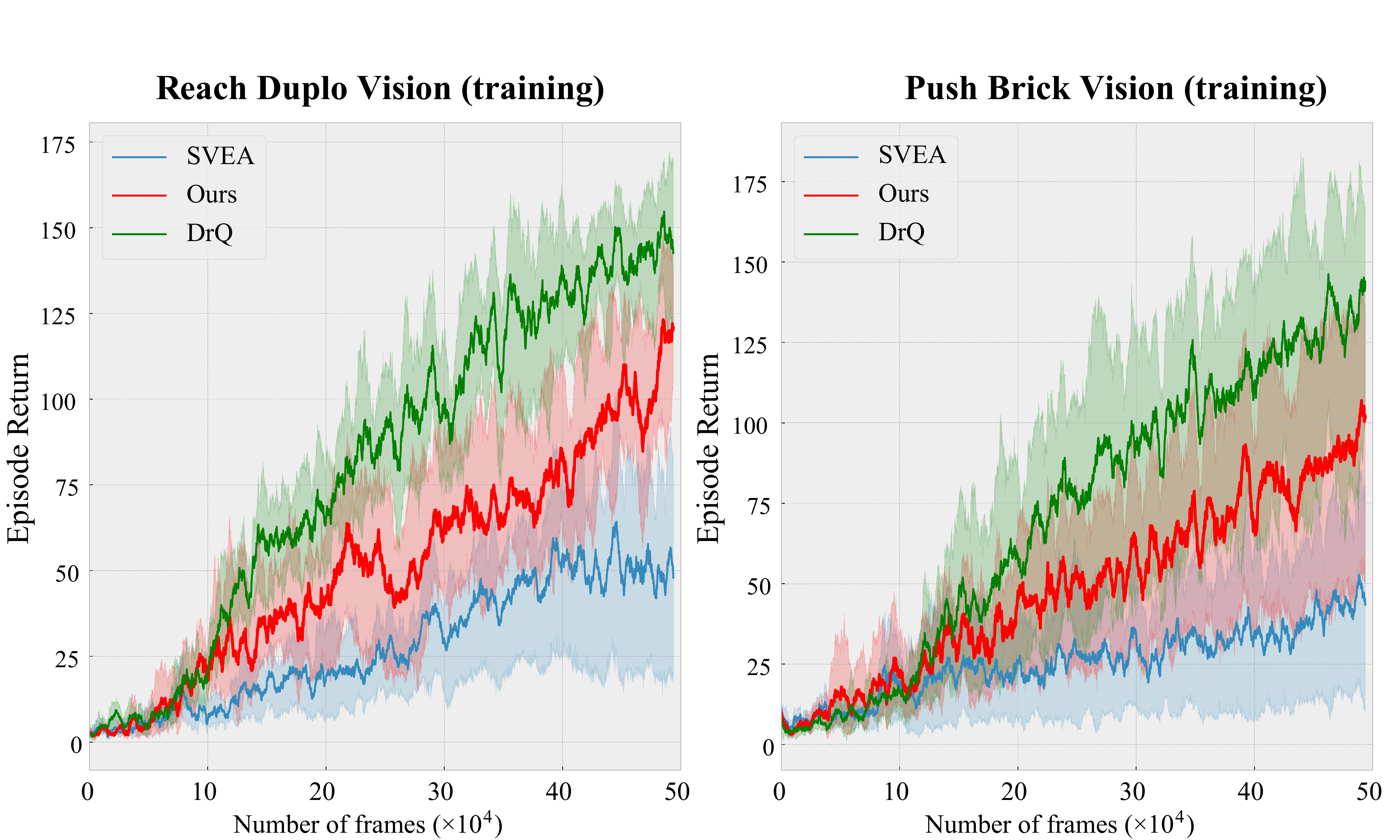}
   \caption{\textbf{Training Performance in Manipulation Tasks}. DrQ (\textit{green line}) shows the best sample efficiency during training. Under strong augmentation, SVEA (\textit{blue line}) may suffer from training divergence while TLDA (\textit{red line}) can still maintain stability.}
   \label{fig7}
\end{figure}

\section{Additional Result} \label{AR}

\subsection{TLDA in DMC}
\begin{figure}[t]
    \centering
    \subfigure[]
        {\includegraphics[width=0.78\linewidth]{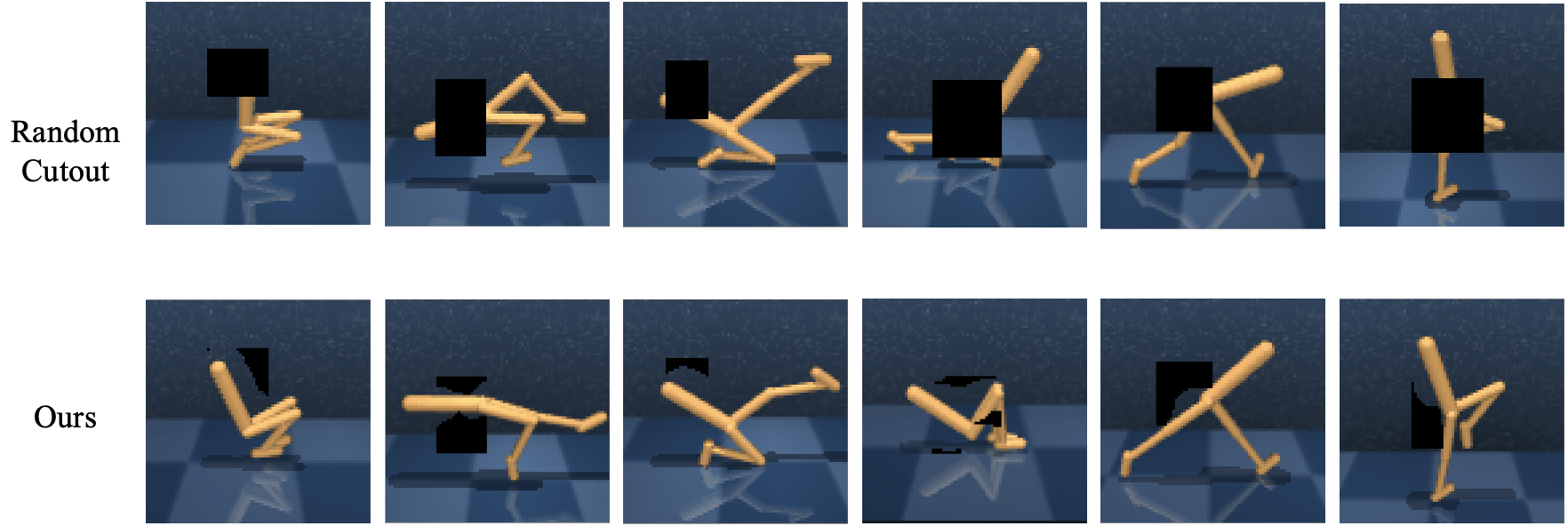}}
    \subfigure[]
        {\includegraphics[width=0.78\linewidth]{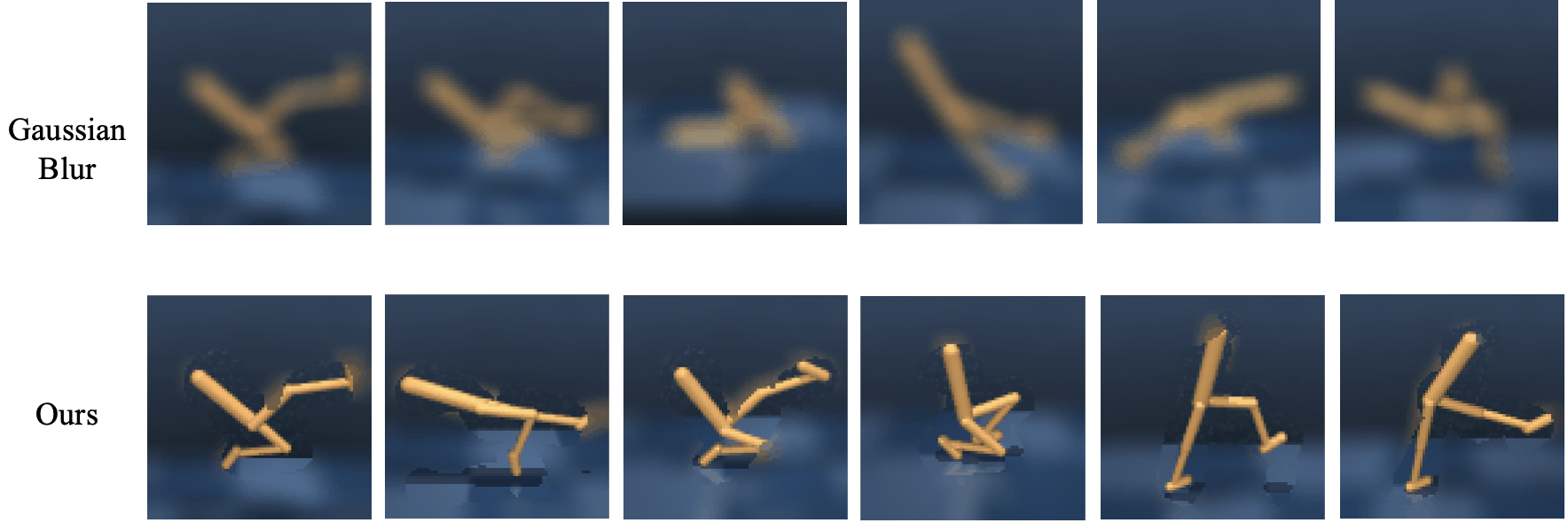}}
    \subfigure[]
        {\includegraphics[width=0.78\linewidth]{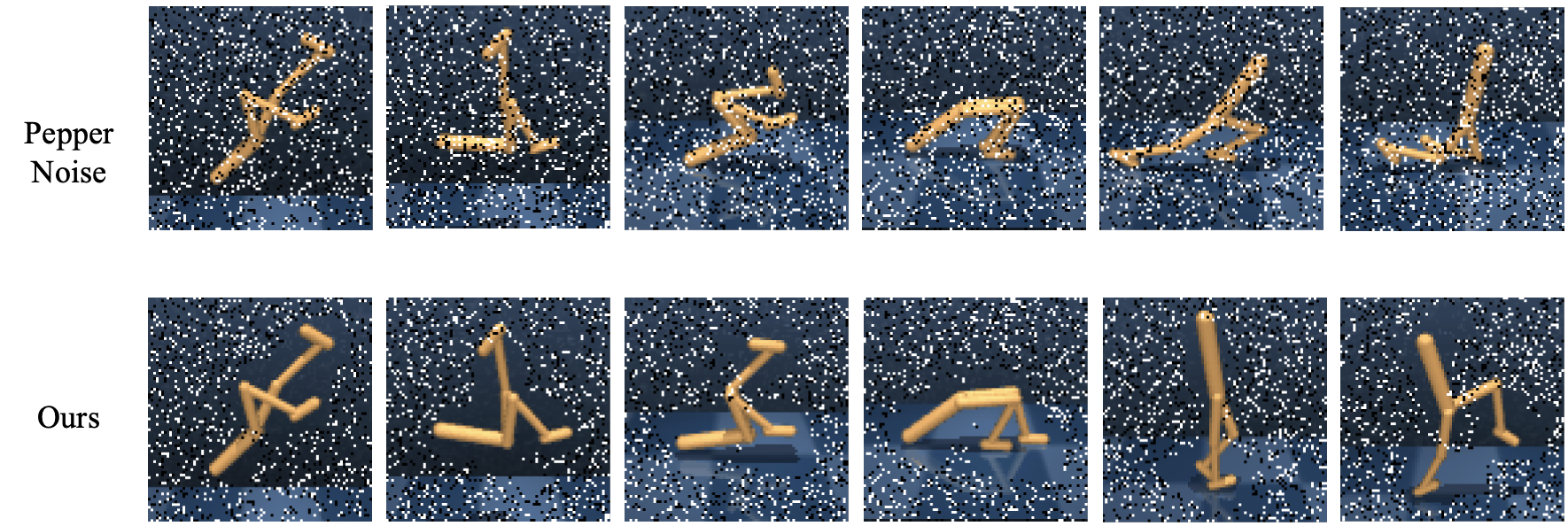}}
    \subfigure[]
        {\includegraphics[width=0.78\linewidth]{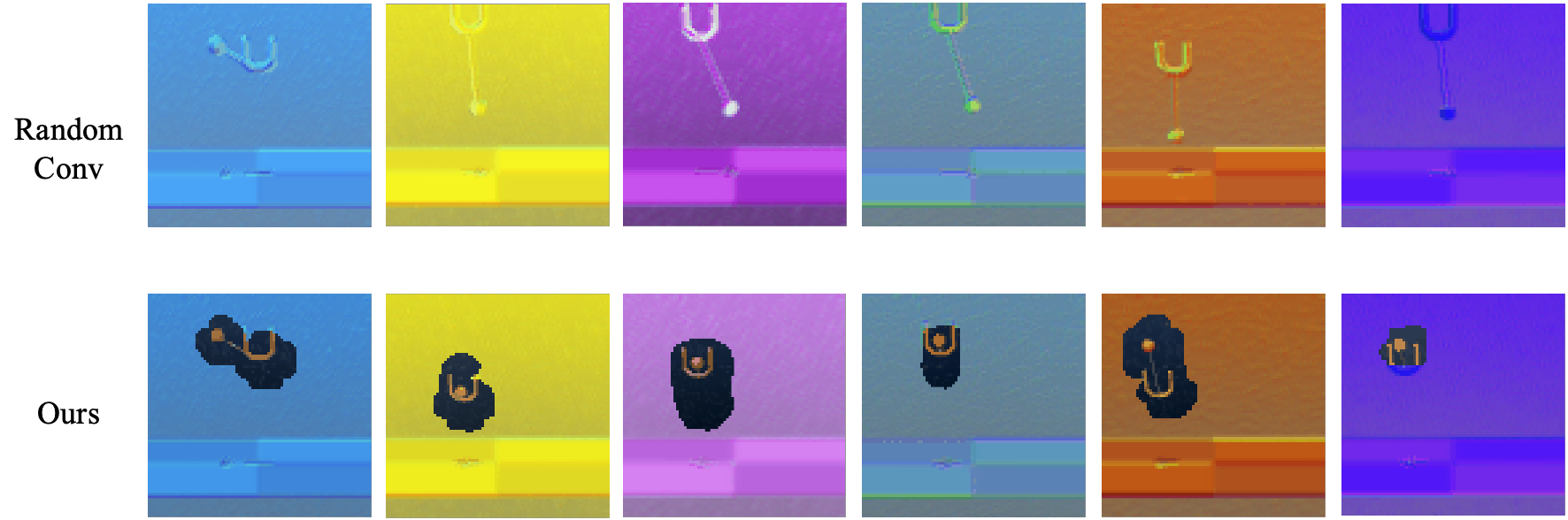}}
\caption{\textbf{TLDA in DM-control-suite.} We compare the converged DrQ agent facing different kinds of augmentations in the same timestep. The results show that TLDA will help the training agent to get a better asymptotic performance.}
\label{fig:more tlda}
\end{figure}

As shown in Figure~\ref{fig:more tlda}, we exhibit more comparisons between the TLDA and normal strong augmentations to show the superiority of our method. We use the same converged DrQ agent under the same seed to evaluate how augmentation methods will influence the agent's performance. More details are in this \colorbox{yellow}{\textbf{link}}\footnote{\url{https://sites.google.com/view/algotlda/home}}. TLDA can effectively help training agents to alleviate the degradation of performance while facing the strong augmentation.

\subsection{Training Curves of Manipulation Tasks.} \label{mani:result}

All methods are trained in 500k steps on Manipulation Tasks, and the training results are shown in Figure~\ref{fig7}. Although DrQ shows the best training performance, comparison results for generalization in Table~\ref{tmani} demonstrate that our method significantly outperforms baselines, which means DrQ may tend to overfit the environment, and SVEA may easily suffer from divergence under these more challenging benchmarks.

\subsection{CARLA} \label{carla}
We adapt the reward function and train agents under the weather whose setting is the same as the previous work~\cite{zhang2020learning}. The training results are shown in Figure~\ref{fig:carla}. The results indicate that TLDA outperforms other baselines in sample efficiency.

Except for the success rate of reaching 100m distance, the crash intensity is also an essential driving metric. We evaluate the crash intensity under different weather and report the average value in the training and unseen environments in Table~\ref{carlaintens}.  The results show that TLDA has the lowest crash intensity compared to the other baselines in unseen environments.

\begin{table}[ht] 
\caption{\textbf{CARLA about Crash Intensity}. We report the average crash intensity under different unseen environments with dynamic weather conditions across 5 seeds runs. The lower crash intensity value indicates the better agent's driving stability.}
\centering
\begin{tabular}{cccc}
\toprule[0.5mm]
Setting         & DrQ   & SVEA  & Ours  \\ \midrule[0.3mm]
Training        & $\bm{1520}$ & $1834$  & $1862$  \\
Wet Noon        & ${2550}$   & $2380$  & $\bm{1524}$   \\
SoftRain noon  & $2463$  & $1300$   & $\bm{975}$   \\
Wet Sunset     &    {$1893$ }  & {$1227$ }   & {$\bm{1160}$}  \\
MidRain Sunset    & {$1808$ }      & {$1561$ }  & {$\bm{767}$}  \\ \hline
\end{tabular}
\label{carlaintens}
\end{table}

\end{document}